\documentclass{article}

\PassOptionsToPackage{numbers}{natbib}
\usepackage[preprint]{neurips_2026}
\usepackage{etoc}
\usepackage{lipsum}
\usepackage{minitoc} 

\usepackage{hyperref}
\usepackage{url}
\usepackage{algorithm}
\usepackage{xcolor}
\usepackage{enumitem}
\usepackage{wrapfig}
\usepackage{makecell}
\usepackage{multirow}

\usepackage{algpseudocode}
\usepackage{booktabs}
\usepackage{multirow} 
\usepackage[table]{xcolor}
\usepackage{arydshln}
\usepackage{makecell}
\usepackage{graphicx}   
\usepackage{subcaption} 
\usepackage{float}

\usepackage{tcolorbox}
\usepackage{caption}
\usepackage{tcolorbox}
\usepackage{amssymb}
\usepackage{mathtools}

\usepackage[utf8]{inputenc} 
\usepackage[T1]{fontenc}    
\usepackage{hyperref}       
\usepackage{url}            
\usepackage{booktabs}       
\usepackage{amsfonts}       
\usepackage{nicefrac}       
\usepackage{microtype}      
\usepackage{xcolor}         

\title{Constraints-of-Thought: A Framework for Constrained Reasoning in Language-Model-Guided Search}

\author{\textbf{Kamel Alrashedy, Vriksha Srihari, Zulfiqar Zaidi, }\\ 
\textbf{Ridam Srivastava, Pradyumna Tambwekar, Matthew Gombolay} \\
  Georgia Institute of Technology, GA, USA \\ 
  \texttt{\{kalrashedy3,vriksha.srihari,zzaidi8\}}@gatech.edu \\
  \texttt{\{rsrivastava7,ptambwekar3\}}@gatech.edu \\
  \texttt{\{matthew.gombolay\}}@cc.gatech.edu}

\begin{document}

\maketitle

\begin{abstract}
  While large language models (LLMs) can generate multi-step plans, they frequently produce steps that violate high-level intent or domain constraints. We introduce Constraints-of-Thought (Const-o-T), a structured intermediate representation that converts each reasoning step into an ⟨intent, constraint⟩ pair, where the constraint is an executable predicate evaluated by a domain validator. These constraints restrict the feasible action space at each state and transform free-form reasoning traces into controllers that actively shape search. \textcolor{black}{The primary contribution of this work is the executable intent–constraint representation, which converts reasoning steps into constraints that directly restrict the planner’s feasible action space. We demonstrate the utility of this representation by integrating it with Monte Carlo Tree Search (MCTS), where extracted constraints restrict the planner’s admissible actions during search.} This yields both improved task accuracy and reduced search complexity, reflected in lower branching factors and faster convergence. Across four domains representing distinct reasoning settings including planning, code synthesis, symbolic reasoning, and scientific knowledge, Const-o-T consistently outperforms prompting-only and search-based baselines. \textcolor{black}{Each reasoning step produces an executable constraint $\kappa_t$ that induces a restricted action set $A_{\kappa_t}(s) \subseteq A(s)$, which is enforced prior to node expansion during search.}The key intellectual contribution is demonstrating that executable intent–constraint representations can serve as a general, \textcolor{black}{domain-agnostic} prior that converts language model reasoning from descriptive rationales into verifiable search controllers.
\end{abstract}

\section{Introduction}

\textcolor{black}{Large language models (LLMs) can generate multi-step reasoning traces, but these traces often violate domain constraints and fail to align with user intent. While approaches such as Chain-of-Thought (CoT) reasoning~\cite{wei2022chain} appear to “think through” the space of possible plans, they produce free-form reasoning without an executable mechanism to enforce feasibility during planning. As a result, generated actions frequently violate domain constraints or logical consistency across a range of tasks, including combinatorial planning (e.g., strategy games and program synthesis~\cite{silver2017mastering, guan2024richelieu, madaan2022language}) and symbolic reasoning (e.g., arithmetic~\cite{cobbe2021training}). Prior work typically uses reasoning traces as guidance or post-hoc verification, but does not modify the planner’s feasible action space, leading to exploration of invalid or misaligned actions during search~\cite{zhou2024can}.}


\textcolor{black}{We propose Constraints-of-Thought (Const-o-T), a framework that converts reasoning steps into executable constraints that restrict the admissible action set during search. Concretely, each constraint defines a predicate $\kappa_t$ over actions that induces a restricted action set $A_{\kappa_t}(s) \subseteq A(s)$, and this restriction is enforced before node expansion during search. Unlike prior approaches that use constraints for scoring, prompting, or post-hoc filtering, which still operate over the full action set $A(s)$, Const-o-T modifies the planner itself by restricting $A(s)$ to $A_{\kappa_t}(s)$. This transformation turns reasoning from descriptive rationales into search controllers, reducing the branching factor and focusing exploration on feasible trajectories.}

\begin{figure*}[t]
  \centering
  \vspace{-55pt}
  \includegraphics[height=6.5cm]{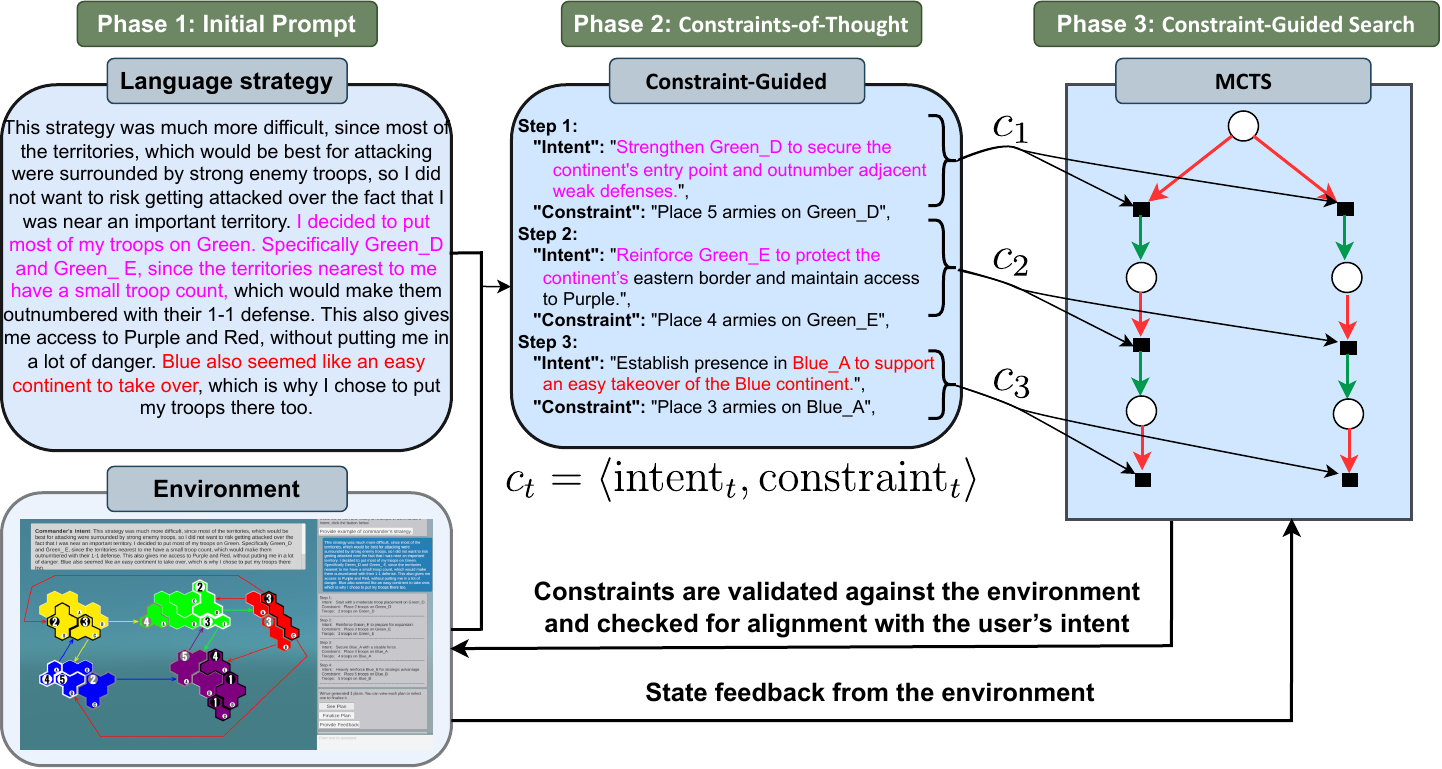}
  \caption{\textcolor{black}{Const-o-T empowers LLMs to (i) infer intent statements and (ii) extract executable constraints from high-level strategies, which restrict the admissible action set during search and act as search controllers rather than post-hoc output filters. These constraints guide a budgeted, constraint-guided variant of MCTS, steering exploration toward rule-compliant and high-quality actions while remaining robust to moderate constraint errors (See example from CAD code Fig \ref{fig:schematicCAD}).}}
  \vspace{-15pt}
  \label{fig:schematic}
\end{figure*}


We demonstrate the practical benefits of this representation by integrating Const-o-T with MCTS for constraint-guided planning. The key novelty of our approach is the executable intent–constraint representation, while MCTS serves as a supporting mechanism that operationalizes these constraints during search. In our framework, an LLM first generates a sequence of ⟨intent, constraint⟩ pairs from the user’s strategic input. Each constraint $\kappa_t$ induces a restricted action set $A_{\kappa_t}(s) \subseteq A(s)$, and this restriction is enforced before node expansion during search, pruning infeasible actions and reducing the branching factor. \textcolor{black}{Prior approaches incorporate constraints through scoring, verification, or affordance weighting, where the full action set $A(s)$ is still considered during generation. In contrast, our approach restricts the admissible action set to $A_{\kappa_t}(s)$ prior to expansion, preventing infeasible actions from entering the search process.} We evaluate Const-o-T across four domains, where it consistently outperforms strong baselines. These results demonstrate that structured intermediate representations can effectively guide planning and improve reliability. Because Const-o-T produces interpretable intent–constraint pairs that expose the system’s reasoning process, we also evaluate whether this representation improves human understanding and perceived alignment with system behavior, finding that users report significantly higher transparency, usability, trust, and alignment. Our work makes the following contributions:
\begin{enumerate}[noitemsep, leftmargin=*]
\item  \textcolor{black}{\textbf{Constraints-of-Thought (Const-o-T).} We propose an executable intermediate representation that decomposes strategies into ⟨intent, constraint⟩ pairs, where constraints define a reduced feasible action space $A'(s)$.}
\item  \textbf{Constraint-Guided MCTS.} We integrate Const-o-T with MCTS, operationalizing executable constraints as search controllers that restrict the feasible action space during exploration.
\item \textbf{Computational Evaluation.} We empirically show consistent branching-factor reduction and compute-normalized gains across heterogeneous domains.
\textcolor{black}{\item \textbf{Human Evaluation.} We conduct a user study on the board game Risk showing that the structured $\langle$intent, constraint$\rangle$ representation improves transparency, trust, and alignment with user strategies ($p < 0.05$).}
\end{enumerate}

\vspace{-5pt}
\section{Related Work}
\vspace{-5pt}
\textbf{Planning and Decision Making.} Classical AI planning relies on heuristic search~\citep{pearl1984heuristics}, probabilistic models such as MDPs~\citep{puterman2014markov}, and simulation-based methods like MCTS~\citep{browne2012survey}, but these approaches require explicit domain modeling~\citep{chakraborti2020emerging}. Recent work explores LLMs as flexible planners, leveraging pretrained knowledge to generate reasoning traces and actions in open-ended settings. Examples include enhancing planning with tree search~\citep{light2025strategist}, combining heuristic reasoning with symbolic search~\citep{saha2024system}, and using episodic memory for long-horizon strategies~\citep{zhu2024language}. \textcolor{black}{Prior LLM–search hybrids typically use language models to generate candidate actions or evaluate nodes during search, rather than modifying the planner’s feasible action space.} In contrast, our approach modifies the feasible action set itself by introducing executable constraints that restrict expansions before search proceeds.

\textcolor{black}{\textbf{Reasoning with LLMs.} LLMs improve performance by reasoning before answering. CoT~\citep{wei2022chain} and ToT~\citep{yao2023tree} generate step-by-step traces that aid tasks such as arithmetic~\citep{cobbe2021training}, commonsense~\citep{zhou2020evaluating}, and decision-making~\citep{huang2022language}. However, these reasoning traces function as explanatory rationales rather than executable constraints and therefore neither define a restricted feasible action set $A'(s)$ within the planning process nor enforce constraint satisfaction during planning. Consequently, reasoning steps cannot formally restrict the actions considered during search, allowing plans to drift from the user’s intent in combinatorial spaces.}

\textbf{Constraint-Guided Reasoning.} Constraints have long ensured feasibility in symbolic planning~\citep{russell1995modern}, and recent work extends this idea to LLMs through structured prompting in program synthesis~\citep{austin2021program} and intent-to-constraint translation~\citep{tambwekar2023computational}. Other studies show constraints reduce hallucinations in reinforcement learning and neural planning~\citep{garcia2012safe}, while LLMFP~\citep{hao2024planning}  formalizes prompts into verifiable representations. \textcolor{black}{However, these approaches primarily use constraints as static templates, optimization objectives, or post-hoc verifiers. Unlike prior approaches that translate intent into constraints for optimization or prompting, we treat constraints as executable controllers that dynamically restrict the \textcolor{black}{planner’s} feasible action set during search.} \textcolor{black}{We distinguish three classes of approaches. Constrained decoding enforces token-level restrictions during sequence generation. Neuro-symbolic \citep{feng2025vericot} and affordance-based methods \citep{ahn2022can} (e.g., SayCan) incorporate feasibility through scoring or weighting candidate actions, but still evaluate the full action set $A(s)$. In contrast, our method applies executable constraints during search, restricting the admissible action set to $A_{\kappa_t}(s)$ before expansion.}

\textbf{Constrained Generation and Decoding.}
\textcolor{black}{Most prior work on constrained generation applies constraints during decoding rather than at the planning level. PPL-MCTS \citep{chaffin2022ppl} integrates constraints into generation by guiding MCTS with a learned discriminator; however, these constraints primarily function as evaluative signals that influence path selection rather than explicitly restricting a structured action space. NeuroLogic \citep{lu2022neurologic} introduces a search-based decoding strategy that enforces logical constraints during generation instead of applying them post-hoc, while DOMINO \citep{beurer2024guiding} improves the efficiency of constrained decoding. However, both operate at the token-sequence level. In contrast, Const-o-T applies constraints at the planning level: each executable constraint induces a reduced admissible action set $\mathcal{A}_{\kappa_t}(s)$ before node expansion. Thus, constraints act as search controllers that shape high-level planning, rather than as token-level decoding rules or post-hoc evaluators.}

\begin{algorithm*}
\caption{Constraint-Guided MCTS (CG-MCTS)}
\label{alg:cg-mcts}
\begin{algorithmic}[1]
\Require Initial state $s_0$; Map $M$; Constraint sequence $\mathcal{C}=\big[(\text{intent}_i,\text{constraint}_i)\big]_{i=1}^{K}$;
\Ensure Plan $\pi$ or root action $a^\star$
\Function{CG\_MCTS}{$s_0, M, \mathcal{C}$}
    \State Create root node $v_0$ with state $s_0$; 
    \State $R \gets |\mathcal{C}| \equiv K$ \Comment{Number of rollouts equals number of constraints}
    \For{$k = 1$ \textbf{to} $R$}
        \State $c \gets \mathcal{C}[k]$ \Comment{Use the $k$-th constraint to guide this rollout}
        \State $v \gets v_0$; $\text{path} \gets [\,]$
        \While{\textbf{not} \Call{Terminal}{$v$}} \Comment{Selection}
            \State $a \gets \arg\max\limits_{a \in \mathcal{A}(v)} \Big[
                    Q(v,a)
                    + c_{\mathrm{uct}} \sqrt{\frac{\ln(1+N(v))}{1+N(v,a)}}
                    + \lambda \cdot \log P_{\mathrm{LM}}(a \mid \text{state}, \kappa_k)
                    \Big]$
           
            \State $\text{path} \gets \text{path} \cup \{(v,a)\}$
            \State $v \gets \Call{Child}{v,a}$
        \EndWhile
        \State  $\tilde{\mathcal{A}} \sim \Call{TopK}{P_{LM}(\cdot \mid \mathrm{state}(v), c, D),\, K_{\text{gen}}}$ \Comment{Expansion} 
        \State $\mathcal{A}_{\text{legal}} \gets \{ a \in \tilde{\mathcal{A}} : \kappa_k(a) = 1 \}$ \Comment{Constraints restrict action space before expansion}
        \State  \textbf{if} $\mathcal{A}_{\text{legal}}=\varnothing$ \textbf{then} $\mathcal{A}_{\text{legal}} \gets \{a \in \tilde{\mathcal{A}} : \mathbb{I}_c(a \mid v)=1\}$ \textbf{else keep top }$K_{\text{expand}}$
        \State  For each $a \in \mathcal{A}_{\text{legal}}$:
        \State \hspace{0.5em} $v' \gets \Call{CreateChild}{v,a,\; \mathrm{NextState}(v,a)}$
        \State $v \gets \Call{Evaluation}{v}$ \Comment{Evaluation}
        \State \Call{Backup}{$\text{path},\, V(v)$} \Comment{Backpropagation}
    \EndFor
    \State $\pi' \leftarrow \Call{ExtractBestPath}{v_0}$ \Comment{Construct plan from search tree}
    \State $\pi \leftarrow \Call{FinalValidation}{\pi', C}$ \Comment{Remove unsupported actions}
    \State \Return $\pi$
\EndFunction
\end{algorithmic}
\end{algorithm*}
\vspace{-8pt}
\section{Constraints-of-Thought Framework for Guided Search}\vspace{-8pt}

\textcolor{black}{We study task planning in domains where translating natural language strategies into executable plans remains challenging.} Existing LLM planners \textcolor{black}{typically} lack an executable intermediate representation that restricts the feasible action set $A'(s)$ during search, causing exploration to include invalid or misaligned actions. \textcolor{black}{We introduce Const-o-T, a structured reasoning framework in which constraints restrict the admissible action set during search, acting as search controllers rather than post-hoc output filters. When integrated with MCTS, this yields a budgeted, constraint-guided variant that focuses exploration on feasible, intent-aligned actions. Moreover, the framework remains robust to moderate constraint errors, maintaining stable performance even when constraint extraction is imperfect.}
\vspace{-10pt}
\subsection{Task Planning}
\vspace{-5pt}

\textcolor{black}{We formulate task planning as a finite-horizon Markov Decision Process (MDP) $(\mathcal{S}, \mathcal{A}, T, r, \gamma, H)$, where the agent generates an action sequence over a horizon of length, $H$, to maximize cumulative reward. The state space, $\mathcal{S}$, captures the relevant environment configuration, and the action space, $\mathcal{A}$, contains the actions available to the agent. The transition function, $T$, defines how actions change the state, and the reward function, $r$, evaluates the quality of decisions. A trajectory is defined as $(s_0, a_0, s_1, a_1, \dots, s_{H-1}, a_{H-1})$, where each action is selected to optimize the expected return.}

\textcolor{black}{The transition dynamics $T$ determine how states evolve in response to actions and may be stochastic or deterministic depending on the domain. The reward function $r$ evaluates the utility of generated outputs, which can be computed from domain-specific fitness functions or external evaluators. In CAD and arithmetic domains, $r$ is approximated using LLM-based judgments of correctness. Importantly, these LLM evaluations provide soft scoring among already feasible candidates, while feasibility itself is enforced by the symbolic constraints used during search. The discount factor $\gamma$ balances immediate and future rewards.} Rather than solving for an optimal policy, $\pi^*$, the planner constructs an action sequence $(a_0, \ldots, a_{H-1})$ that maximizes cumulative reward given by Eq.~\ref{eq:r}-\ref{eq:F}.
\begin{align}
R_{\text{total}} &= \sum_{t=0}^{H-1} \gamma^t r(s_t, a_t) \label{eq:r} \\
r(s,a) &= \mathbf{1}\{\text{constraint satisfied}\} + F(s,a) \label{eq:r_little} \\
F(s,a) &=
\begin{cases}
z_1(s,a), & \text{Risk} \\
z_n(s,a), & \text{Other tasks}
\end{cases}
\label{eq:F}
\end{align}

In these equations, $z_1$ denotes the Risk fitness function (Sec.~\ref{Fitness_function}), while $z_2, \dots, z_n$ represent task-specific evaluation functions for other domains using LLM-as-a-Judge scoring mechanisms. \textcolor{black}{The indicator function $\mathbf{1}\{\cdot\}$ enforces constraint satisfaction by excluding infeasible actions, while $F(s,a)$ provides a task-specific evaluation signal used to rank feasible candidates.}

\vspace{-5pt}
\subsection{Constraints-of-Thought}
\vspace{-5pt}
We introduce \textit{Constraints-of-Thought} (Const-o-T), a framework that represents structured reasoning steps guiding the agent’s decision-making process. \textcolor{black}{Const-o-T represents each reasoning step as an intent–constraint pair. 
Formally, a Const-o-T step is defined as $c_t = \langle i_t, \kappa_t \rangle$, 
where $i_t$ is a natural-language intent describing the strategic objective 
(e.g., ``Reinforce a border to deter enemy") and $\kappa_t$ is a machine-executable 
constraint that enforces the corresponding action restriction 
(e.g., ``Place 5 troops on Territory A").} \textcolor{black}{Once extracted, each $c_t = \langle i_t, \kappa_t \rangle$ pair serves a dual role in planning.} The intent provides a human-interpretable explanation of the agent’s reasoning, while the constraint restricts the feasible action space by pruning actions that violate domain rules. Formally, given a state $s_t$, the constraint defines a reduced action set, $A'(s_t) \subseteq A(s_t)$, containing only actions consistent with $\kappa_t$. This conversion of intermediate reasoning into executable action-set restrictions constitutes the core representation introduced by Const-o-T. Unlike prior LLM–search hybrids that use language models to generate candidate actions or evaluate nodes, Const-o-T uses intermediate reasoning to explicitly modify the planner’s feasible action set. In this formulation, reasoning steps act as executable controllers that restrict the search space rather than merely guiding exploration. \textcolor{black}{The role of symbolic validation is to enforce feasibility constraints during search by preventing invalid actions from entering the search tree. Candidate actions that do not satisfy $\kappa_t$ are rejected prior to expansion. Constraints follow simple domain-specific schemas (e.g., ‘Place n troops on territory x’) that map to executable predicates $\kappa_t$. Validation is performed using domain-specific rule checkers (e.g., legality constraints in Risk, compilation checks in CAD, and symbolic consistency in arithmetic reasoning)}

\vspace{-5pt}
\subsection{Constraint-Guided MCTS}
\vspace{-5pt}

\textcolor{black}{We employ a constraint-guided, budgeted variant of MCTS. Unlike classical MCTS, which targets asymptotic optimality under large rollout budgets, our objective is efficient, intent-aligned search under limited compute.}

\textbf{Algorithm Overview:} Given a natural language strategy, we first extract a sequence of $\langle i_t, \kappa_t \rangle$ pairs using a language model (See Appendix~\ref{App_const}). \textcolor{black}{Each constraint $\kappa_t$ defines a predicate over actions that induces a restricted action set $A_{\kappa_t}(s) \subseteq A(s)$, which is enforced before node expansion during search (See Algorithm~\ref{alg:cg-mcts}). The LLM proposes constraints, while the symbolic validator enforces feasibility by filtering invalid actions before they enter the search tree.}



\textbf{Selection Phase:} \textcolor{black}{Starting from the root, MCTS recursively selects child nodes using a modified Upper Confidence Bound (UCB) criterion that balances exploration and exploitation with the LLM's confidence, as given by Eq.~\ref{eq:UCB}.}
\begin{equation}
\text{UCB}(v,a) = Q(v,a) + c_{\text{uct}} \sqrt{\frac{\ln(1+N(v))}{1+N(v,a)}} + \lambda \cdot \log P_{\text{LM}}(a \mid \text{state}, \kappa_t)
\label{eq:UCB}
\end{equation}
\textcolor{black}{In this equation, $P_{\text{LM}}(a \mid \text{state}, \text{$\kappa_t$})$ denotes the probability assigned by the language model to action $a$ given the current state and constraint, and $\lambda$ controls the influence of this prior during search.}



\textbf{Constraint-Guided Search.} \textcolor{black}{Const-o-T restricts the feasible action set during MCTS by enforcing constraint predicates at expansion time, pruning actions that violate the extracted intent–constraint pairs. This reduces branching factor and focuses search on strategies consistent with user intent.}

\textbf{Expansion and Evaluation.} At leaf nodes, LLMs generate candidate actions given the current state, active constraint, and user strategy. We filter candidates for legality (e.g., game rules) and constraint satisfaction before expansion. Each resulting child node is evaluated using a domain-specific fitness function measuring strategic objective achievement and constraint adherence. \textcolor{black}{The rollout budget is set equal to the number of extracted constraints, coupling search depth to the structured constraint sequence.}

\vspace{-5pt}
\paragraph{Constraint-Guided Reasoning.}
\textcolor{black}{
We formalize constraint-guided reasoning as a structured planning process in which intermediate reasoning steps restrict the feasible action space of a planning algorithm. Let $S$ denote the state space and let $\mathcal{A}(s)$ denote the set of admissible actions in state $s$.
Each Const-o-T reasoning step $c_t=\langle i_t,\kappa_t\rangle$ restricts the feasible action space of the planner, where $i_t$ denotes a natural-language intent and $\kappa_t$ denotes an executable constraint. The constraint predicate is defined as shown in Eq.~\ref{eq:constraint_predicate}.} In this equation, $\kappa_t(a)=1$ when action $a$ satisfies the extracted constraint at state $s_t$, and $\kappa_t(a)=0$ otherwise.
\vspace{-3pt}
\begin{equation}
\kappa_t : \mathcal{A}(s_t) \rightarrow \{0,1\}
\label{eq:constraint_predicate}
\end{equation}

Constraints influence planning through two complementary mechanisms. First, during search, the LLM conditioned on $\kappa_t$ guides MCTS toward constraint-consistent candidate actions. At expansion time, candidate actions are checked against $\kappa_t$, and actions that do not satisfy the active constraint are not added to the search tree. Second, after a full plan is generated, a domain-specific validator checks the final output and removes unsupported actions. This final validation step is conservative: it does not introduce new actions, regenerate the plan, or iteratively repair the output.

The constrained action space induced by the predicate is given in Eq.~\ref{eq:restricted_action_space}.
\begin{equation}
\mathcal{A}_{\kappa_t}(s_t)=\{a\in \mathcal{A}(s_t)\mid \kappa_t(a)=1\}
\label{eq:restricted_action_space}
\end{equation}
Given a sequence of constraints $C=(\kappa_1,\dots,\kappa_T)$, the planner generates an action trajectory $\tau=(a_1,\dots,a_T)$ such that $a_t\in \mathcal{A}_{\kappa_t}(s_t)$ for each step $t$.


Validators are domain-specific and enforce task-level feasibility. In the Risk domain, the final validation step is implemented as a Python rule-based checker that takes as input the generated plan and the extracted intent–constraint pairs and identifies actions that violate the constraints. In CAD generation and arithmetic reasoning, we use LLM-based validators to check that generated outputs satisfy geometric, symbolic, and constraint consistency requirements. Because final validation only removes unsupported actions, it may produce a partial plan when later actions depend on earlier ones that are removed. In practice, we prefer removing known invalid actions and deferring repair to downstream processes or human oversight, rather than presenting infeasible actions as valid.

This procedure contrasts with CoT reasoning, which produces free-form rationales that guide reasoning implicitly but do not modify the planner’s feasible action set. In contrast, Const-o-T produces structured reasoning controllers $c_t=f_{\text{LLM}}(s_t)=\langle i_t,\kappa_t\rangle$, where the constraint $\kappa_t$ explicitly modifies the action space used during search.

\vspace{-5pt}
\section{Experiments}
\vspace{-8pt}
\textcolor{black}{We evaluate our approach along three dimensions: task accuracy, structural alignment of generated plans, and interpretability of intermediate reasoning steps.
We benchmark our} method against a diverse set of baselines spanning prompting-based, search-based, and constraint-based planning approaches. The prompting baselines include Direct Prompt, Chain-of-Thought (CoT) \citep{wei2022chain}, Tree-of-Thought (ToT) \citep{yao2023tree}, CoT with Rejection Sampling (CoT+RS) \citep{yao2025optimizing}, Constraint Template Prompting (CTP) \citep{liu2023pre}, and Chain-of-Verification (CoV) \citep{dhuliawala2024chain}. For search-based methods, we consider classical MCTS \citep{zhang2023planning} as well as MCTS augmented with CoT reasoning. We also include a constraint-based planning baseline, LLMFP \citep{hao2024planning}, which translates natural language problems into formal optimization constraints. Unless otherwise specified, all baselines are evaluated consistently across all domains.

\vspace{-5pt}
\subsection{Game Strategy: Risk}
\vspace{-8pt}

Risk is a strategy board game played on a world map of 21 territories grouped into six continents. We focus on the initial troop placement studied in prior work~\citep{tambwekar2023computational} (See Appendix $\S$\ref{CI_examples} for illustrative example). We use the Commander's Intent (CI) dataset~\citep{tambwekar2023computational}, which contains $1,053$ examples of natural language strategies paired with corresponding ground-truth territories for troop placement. We measure alignment using Earth Mover’s Distance (EMD)~\cite{rubner2000earth}, which quantifies the minimum cost of \textcolor{black}{transforming the predicted troop placement distribution into the ground-truth distribution.} Our results demonstrate that across all evaluated LLMs, MCTS with Const-o-T consistently achieves the lowest EMD, indicating the best alignment between predicted and ground-truth troop placements.
The key improvement comes from constraint-guided search, which restricts the action space to strategy-consistent troop placements during planning (See Table~\ref{tab:risk_game}).

\begin{table*}
\centering
\caption{EMD between predicted and ground-truth troop placements across multiple LLMs. Lower values indicate better spatial alignment with human strategies.}

\resizebox{0.95\textwidth}{!}{%
\begin{tabular}{>{\raggedright\arraybackslash}p{5.2cm}
                |c c c c c c}
\toprule
\textbf{Method} 
& \textbf{LLaMA 3.3}  
& \textbf{GPT-4.1} 
& \textbf{GPT-5.2} 
& \textbf{Claude Haiku} 
& \textbf{DeepSeek R1} 
& \textbf{GPT-OSS-120B}  \\
\midrule

Direct Prompt
& 0.64
& 0.60
& 0.58
& 0.63
& 0.76
& 0.61 \\

CoT~\citep{wei2022chain}
& 0.74
& 0.54
& 0.56
& 0.64
& 0.73
& 0.70 \\

CoT+RS~\citep{yao2025optimizing}
& 0.72
& 0.64
& 0.70
& 0.70
& 0.87
& 0.66 \\

LLMFP~\cite{hao2024planning}
& 1.13
& 0.81
& 1.08
& 1.03
& 1.15
& 0.82 \\

CTP~\citep{liu2023pre}
& 0.68
& 0.56
& 0.59
& 0.65
& 0.85
& 0.60 \\

CoV~\citep{dhuliawala2024chain}
& 0.70
& 0.56
& 0.59
& 0.63
& 0.79
& 0.62 \\

Const-o-T
& 0.66
& 0.60
& 0.61
& 0.65
& 0.77
& 0.64 \\

ToT~\cite{yao2023tree}
& 1.05
& 0.72
& 0.76
& 0.89
& 0.90
& 0.85 \\

MCTS~\citep{zhang2023planning}
& 0.96
& 0.76
& 0.79
& 0.79
& 1.01
& 0.93 \\

MCTS with CoT
& 0.99
& 0.75
& 0.70
& 0.76
& 1.01
& 0.82 \\

MCTS with Const-o-T
& \colorbox{green!25}{\textbf{0.57}}
& \colorbox{green!25}{\textbf{0.51}}
& \colorbox{green!25}{\textbf{0.54}}
& \colorbox{green!25}{\textbf{0.55}}
& \colorbox{green!25}{\textbf{0.59}}
& \colorbox{green!25}{\textbf{0.59}} \\

\bottomrule
\end{tabular}
}

\label{tab:risk_game}
\end{table*}

\begin{figure*}
  \centering
  \includegraphics[width=\linewidth]{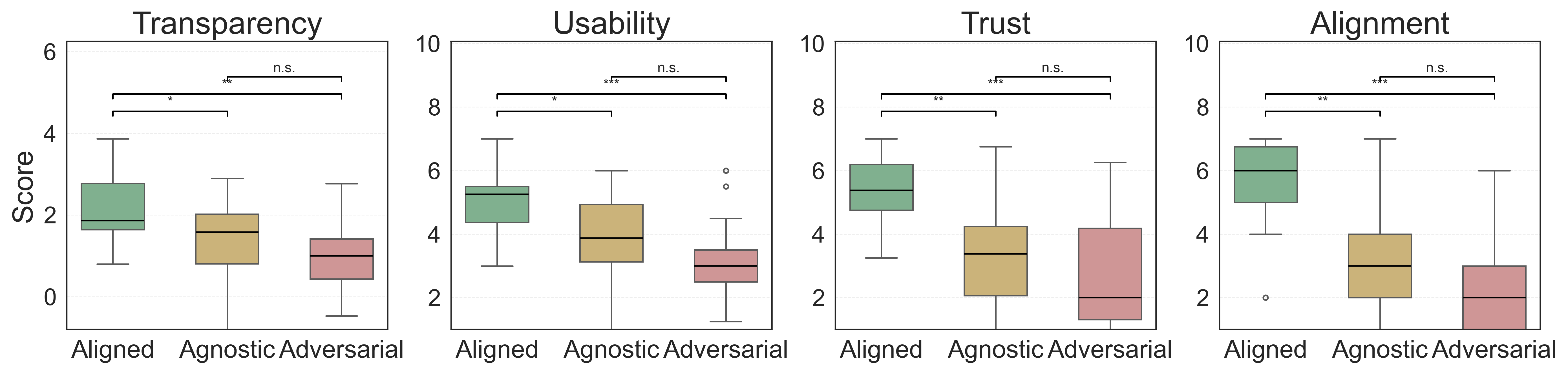}
    \caption{User study ratings across three interaction modes: alignment, agnostic, and adversarial. Statistical significance is indicated by asterisks ($^*p < 0.05$, $^{**}p < 0.01$, $^{***}p < 0.001$).}

  \label{fig:user_study}
\end{figure*}

\begin{table*}[t]
\centering
\caption{Evaluation metrics across benchmarks. For CADPrompt, we report the median Hausdorff Distance (HD) and Success Rate (SR). For GSM8K, Math-500, and GPQA, we report accuracy.}

\footnotesize
\setlength{\tabcolsep}{3pt}
\label{res:CADcode_Math_All}
\resizebox{\textwidth}{!}{
\begin{tabular}{>{\raggedright\arraybackslash}p{3.0cm}
                c c | c c | c c | c c | c c}
\toprule
\multirow{3}{*}{\textbf{Method}} 
& \multicolumn{4}{c|}{\textbf{CADPrompt \citep{alrashedy2024generating}}}
& \multicolumn{2}{c|}{\textbf{GSM8K \citep{cobbe2021training}}}
& \multicolumn{2}{c|}{\textbf{Math-500 \citep{cobbe2021training}}}
& \multicolumn{2}{c}{\textbf{GPQA\citep{rein2024gpqa} }} \\
\cmidrule(lr){2-5} \cmidrule(lr){6-7} \cmidrule(lr){8-9} \cmidrule(lr){10-11}
& \multicolumn{2}{c}{\textbf{LLaMA-3}  } 
& \multicolumn{2}{c|}{\textbf{GPT-4.1}  }
& \textbf{LLaMA-3  } & \textbf{GPT-4.1}  
& \textbf{LLaMA-3  } & \textbf{GPT-4.1}  
& \textbf{LLaMA-3  } & \textbf{GPT-4.1}   \\
\cmidrule(lr){2-3} \cmidrule(lr){4-5}
\cmidrule(lr){6-6} \cmidrule(lr){7-7}
\cmidrule(lr){8-8} \cmidrule(lr){9-9}
\cmidrule(lr){10-10} \cmidrule(lr){11-11}
& \textbf{HD$\downarrow$} & \textbf{SR$\uparrow$} 
& \textbf{HD$\downarrow$} & \textbf{SR$\uparrow$} 
& \textbf{Acc.} & \textbf{Acc.}
& \textbf{Acc.} & \textbf{Acc.}
& \textbf{Acc.} & \textbf{Acc.} \\
\midrule
CoT   
& 0.22 & 88.5\%
& 0.15 & 92.0\%
& 91.9\% & 95.1\%
& 64.2\% & 79.2\%
& 36.3\% & 65.9\% \\

CoT + RS  
& 0.27 & 91.5\%
& 0.13 & 92.5\%
& 92.4\% & 93.1\%
& 66.6\% & 78.4\%
& 40.9\% & 66.6\% \\

CoV 
& 0.27 & 89.0\%
& 0.14 & 94.0\%
& 91.3\% & 93.6\%
& 68.2\% & 81.7\%
& 43.0\% & 65.4\% \\

CTP 
& 0.24 & 89.0\%
& 0.16 & 94.0\%
& 86.8\% & 94.0\%
& 65.0\% & 78.3\%
& 38.5\% & 65.9\% \\

Const-o-T  
& 0.22 & 89.5\%
& 0.16 & 94.5\%
& 92.5\% & 96.1\%
& 64.2\% & 79.2\%
& 42.9\% & 64.1\% \\

ToT  
& 0.22 & \colorbox{green!25}{\textbf{94.0\%}}
& 0.15 & 92.0\%
& 92.6\% & 93.7\%
& 60.4\% & 80.8\%
& 41.9\% & 65.6\% \\

MCTS  
& 0.24 & 89.5\%
& 0.13 & 94.5\%
& 92.9\% & 95.0\%
& 66.2\% & 82.2\%
& 48.4\% & 65.1\% \\

MCTS with CoT  
& 0.24 & 90.5\%
& 0.16 & 93.0\%
& 91.8\% & 95.1\%
& 68.8\% & 80.2\%
& 50.5\% & 67.1\% \\

MCTS with Const-o-T  
& \colorbox{green!25}{\textbf{0.21}} & 92.0\%
& \colorbox{green!25}{\textbf{0.12}} & \colorbox{green!25}{\textbf{95.5\%}}
& \colorbox{green!25}{\textbf{93.4\%}} & \colorbox{green!25}{\textbf{96.2\%}}
& \colorbox{green!25}{\textbf{70.0\%}} & \colorbox{green!25}{\textbf{83.1\%}}
& \colorbox{green!25}{\textbf{52.0\%}} & \colorbox{green!25}{\textbf{70.2\%}} \\
\bottomrule
\end{tabular}
}
\label{res:CADcode_Math}
\end{table*}

\vspace{-5pt}
\subsection{CAD Code Generation}
\vspace{-8pt}

The task involves translating natural language descriptions of 3D objects into executable parametric CAD scripts, where constraint enforcement helps maintain syntactic validity and geometric correctness. We utilize CADPrompt~\cite{alrashedy2024generating}, a dataset containing 200 3D design examples, each paired with a natural language description and a corresponding Python script (See Figure~\ref{fig:CADPromptExample_1}). We report Hausdorff distance between generated and ground-truth objects and the percentage of generated examples compiling successfully. The results in Table~\ref{res:CADcode_Math} demonstrate that MCTS with Const-o-T achieves the best performance across both LLaMA and GPT-4.1, with the lowest Hausdorff distances (0.21 and 0.12) and the highest success rate of 95.5\% for GPT-4.1.

\vspace{-5pt}
\subsection{Arithmetic Reasoning}
\vspace{-8pt}
Arithmetic reasoning involves solving multi-step mathematical word problems by translating natural language into symbolic equations and sequential calculations. The task requires symbolic consistency across steps, with performance typically measured using exact-match accuracy on the final answer. We utilize two benchmarks: (i) GSM8K~\citep{cobbe2021training}, which contains 8.5K math word problems (7.5K train, 1K test); and (ii) Math-500~\citep{hendrycks2021measuring}, a benchmark of 500 problems. The results in Table~\ref{res:CADcode_Math} show that integrating structured reasoning improves accuracy across models. On GSM8K, the best results are achieved by MCTS with Const-o-T, reaching 93.4\% accuracy with LLaMA-3 and 96.2\% with GPT-4.1. On Math-500, LLaMA-3 achieves 70.0\% accuracy, while GPT-4.1 achieves 83.1\%.

\vspace{-5pt}
\subsection{Scientific Reasoning}
\vspace{-8pt}

GPQA consists of graduate-level multiple-choice scientific reasoning questions written by domain experts in biology, physics, and chemistry~\citep{rein2024gpqa}. We evaluate LLM performance on GPQA using accuracy. \textcolor{black}{Table~\ref{res:CADcode_Math} shows that constraint-guided reasoning improves GPQA performance, with MCTS + Const-o-T achieving the best results (52.0\% with LLaMA and 70.2\% with GPT-4.1).}


\vspace{-5pt}
\subsection{Human Evaluation}
\vspace{-8pt}

We additionally evaluate whether the structured intent–constraint representation improves human understanding and perceived alignment with system behavior. We conducted a within-subjects user study ($n=18$) under an IRB-approved protocol in Risk game. Participants provided natural-language descriptions of their \emph{commander's intent}, and the system generated corresponding action plans across full gameplay turns (Reinforce, Attack, and Freemove phases). We consider three interaction modes: (1) \emph{Aligned}, where the system follows the user’s stated strategy; (2) \emph{Agnostic}, where the system optimizes only for winning; and (3) \emph{Adversarial}, where the system intentionally acts against the user’s intent. This design isolates whether alignment is genuinely attributable to the intent–constraint representation. We measured four dependent variables using 7-point Likert scales: Alignment, Transparency~\cite{silva2023explainable}, Usability~\cite{lewis2018system}, and Trust as shown in Fig.~\ref{fig:user_study}. The representation improves human interpretability and alignment. Additional details on study design is provided in Appendix $\S$ \ref{User_Study_app}.

\begin{figure}[t]
\centering
\vspace{-35pt}
\begin{center}
\begin{minipage}{0.49\textwidth}
    \centering
    \includegraphics[width=\linewidth,height = 4cm]{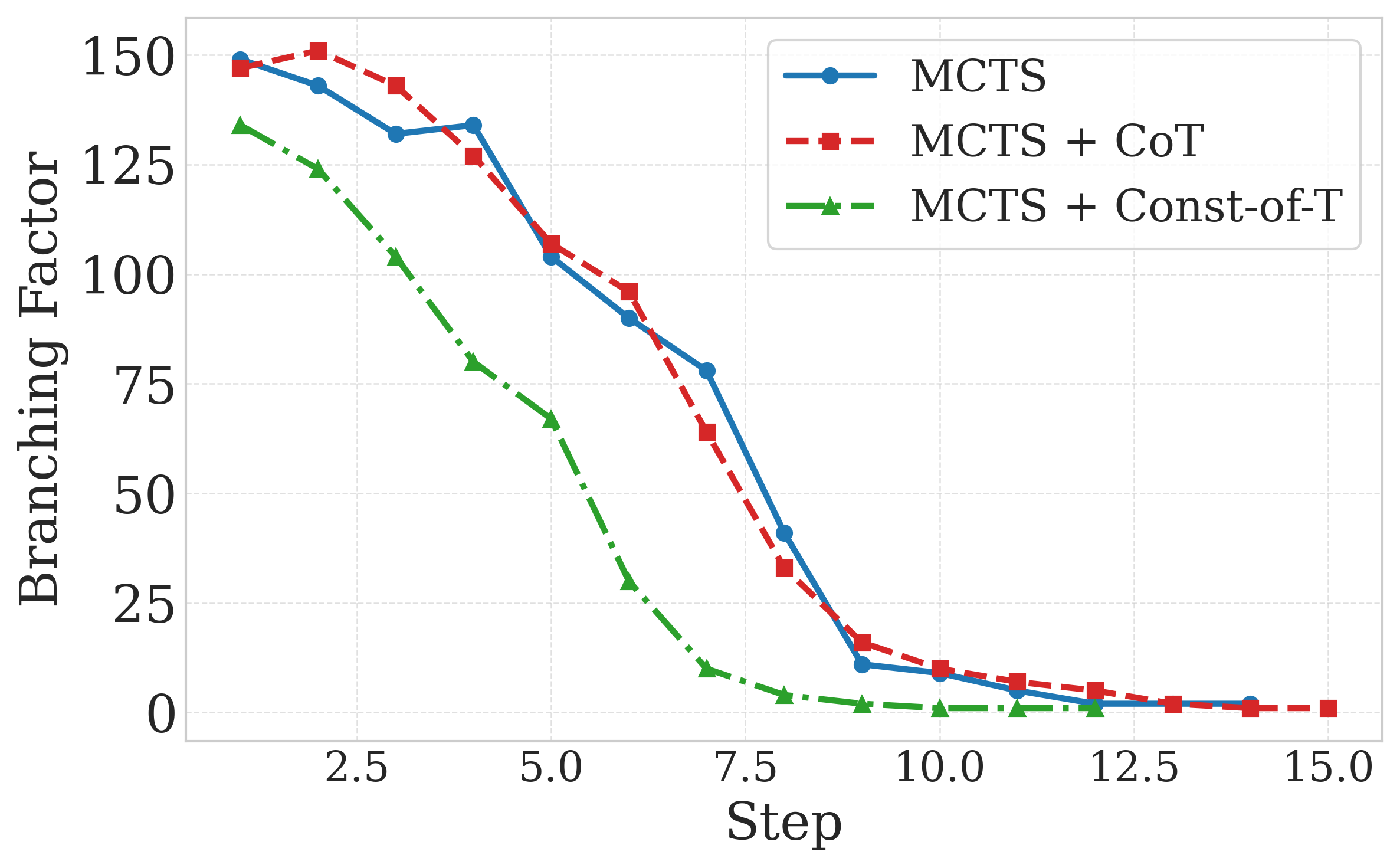}
    \captionof{figure}{Branching factor per step (GPT4.1).}
    \label{fig:branching_factor}
\end{minipage}
\hfill
\begin{minipage}{0.49\textwidth}
    \centering
    \includegraphics[width=\linewidth,height = 4cm]{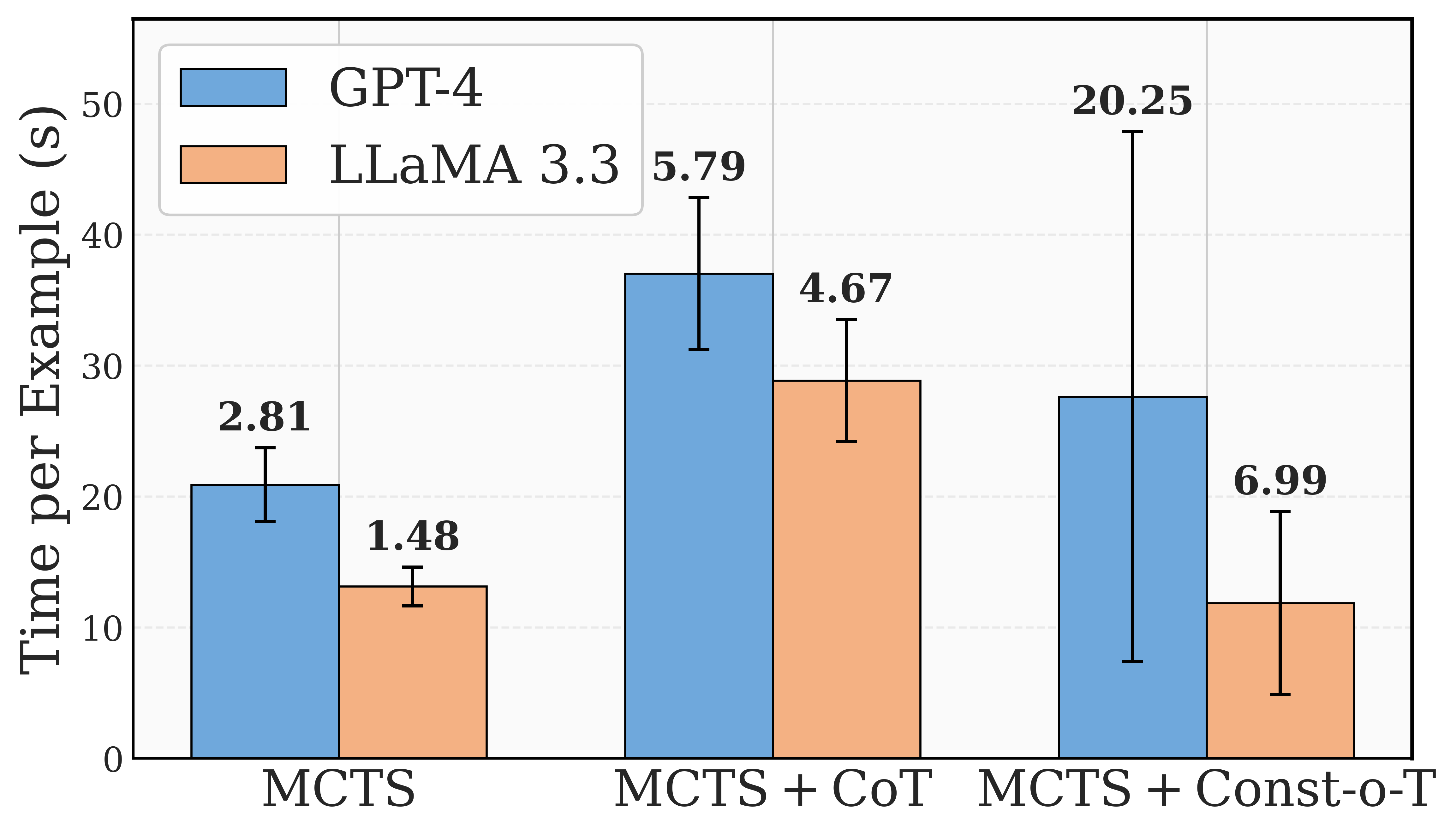}
    \captionof{figure}{Average inference time per example.}
    \label{fig:wallclock}
\end{minipage}
\end{center}
\end{figure}

\section{Analysis}
\vspace{-8pt}
\textcolor{black}{In this section, we analyze our approach in the Risk domain to understand its performance gains. Const-o-T operates through a simple mechanism. Each constraint $\kappa_t$ induces a restricted action set $A_{\kappa_t}(s) \subseteq A(s)$, which is enforced prior to node expansion. This reduces the branching factor by preventing infeasible actions from entering the search tree, leading to more efficient exploration and improved alignment with user intent.}

\label{sec:theory}

    



\textbf{Search-Space Reduction.} \textcolor{black}{A key mechanism underlying the performance gains of our approach is the explicit reduction of the search space. By enforcing constraints during search, Const-o-T restricts the admissible action set prior to node expansion. This leads to a more focused exploration process that avoids infeasible or irrelevant trajectories. Empirically, we observe consistently lower branching factors compared to both standard MCTS and MCTS with CoT (See Fig.~\ref{fig:branching_factor}) on GPT-4.1. \textbf{This confirms that restricting $A_{\kappa_t}(s)$ reduces the search space and improves efficiency.}}



\textbf{\textcolor{black}{Search Efficiency.}}
Figure~\ref{fig:wallclock} shows the average wall-clock time per example. While CoT increases runtime, Const-o-T reduces overall inference time. These improvements shown in Const-o-T arise from restricting the admissible action set during search, which reduces the branching factor and limits the number of node expansions. As a result, the search avoids exploring infeasible trajectories and converges more efficiently. In contrast, CoT introduces additional reasoning overhead without constraining the search space, leading to higher runtime. \textbf{This shows that restricting $A_{\kappa_t}(s)$ improves search efficiency and reduces runtime.}

\begin{wrapfigure}{r}{0.50\linewidth}
\centering
\includegraphics[width=\linewidth]{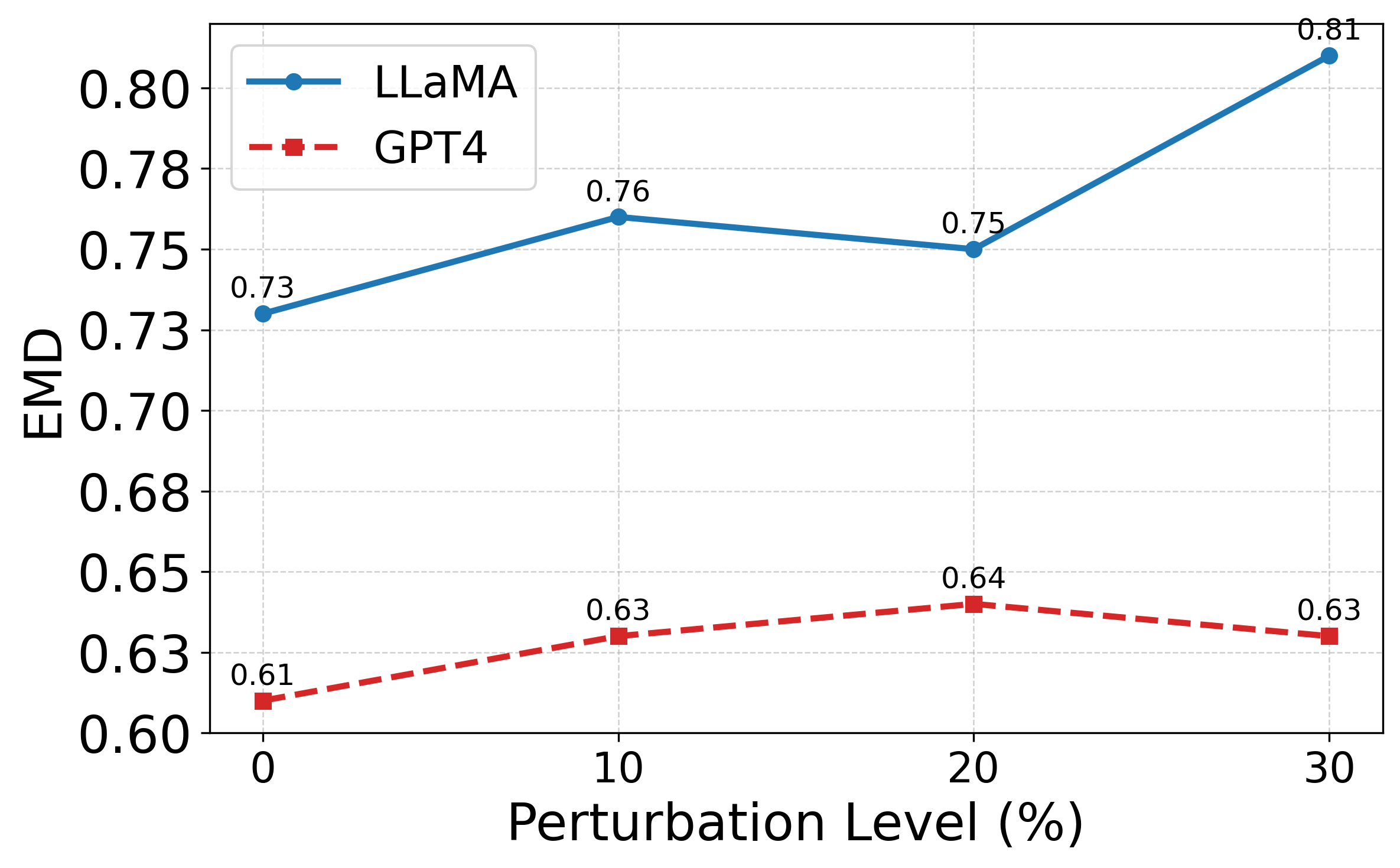}
\caption{EMD vs.\ perturbation level for LLaMA and GPT-4.1 }
\label{fig:corruption_sensitivity}
\end{wrapfigure}

\textbf{Sensitivity to Constraint Quality.}
To evaluate robustness to imperfect constraint extraction, we introduce controlled noise by perturbing a subset of constraints. Specifically, we sample 100 examples from the Risk game using GPT-4.1, where the troop count in selected constraints is modified. Noise is applied at varying rates, $\epsilon \in \{0\%, 10\%, 20\%, 30\%\}$, to simulate increasing levels of extraction error. After constraint extraction, we randomly select $\epsilon$ of the constraints and use an LLM to generate incorrect troop-count values for those entries. We observe gradual performance degradation as noise increases, indicating that the method does not fail catastrophically under imperfect constraints (See Fig.~\ref{fig:corruption_sensitivity}). \textbf{This shows the method remains robust under moderate constraint errors.}

\textbf{In-Search vs Post-hoc Constraints.} \textcolor{black}{We compare our method, which applies executable constraints during search, to a post-hoc filtering baseline in which constraints are applied only after candidate generation, on a subset of 100 examples from Risk using GPT-4.1. All comparisons use identical candidate generation budgets and evaluation settings. Applying constraints during search yields improved performance (from 0.86 to 0.75) and reduces runtime (from 13.6s to 9.8s). In contrast, post-hoc filtering exhibits a high failure rate due to early constraint violations, which prevents effective exploration. This shows that enforcing constraints during search leads to more efficient exploration by avoiding infeasible trajectories early, rather than discarding them after generation. This result supports the claim that constraints define the admissible action set during search rather than acting as post-hoc filters. \textbf{This shows constraints must be applied during search rather than post-hoc.}}

\vspace{-2pt}
\textbf{Failure Modes.} We identify three primary failure modes in the Risk domain. First, errors in constraint extraction can impose infeasible or misaligned restrictions on the action space. Second, overly restrictive constraints can lead to over-pruning, eliminating valid and potentially high-quality actions. Third, ambiguous or underspecified user input may result in incomplete constraint sets (See Fig.~\ref{fig:Failure_Modes}). \textbf{This highlights the primary limitations arising from constraint extraction and over-pruning.}

\vspace{-10pt}
\section{Discussion}
\vspace{-10pt}

\textbf{From Rationales to Controllers.} While CoT produces unconstrained reasoning traces that primarily function as explanatory rationales, Const-o-T represents intermediate reasoning as executable intent–constraint pairs that directly restrict the feasible action set during search. \textcolor{black}{The key distinction is whether constraints define the admissible action set or merely evaluate candidates. Our method explicitly defines the constraint-induced action set $A_{\kappa_t}(s)$, which directly shapes the planner’s exploration by pruning infeasible actions before expansion}. Table~\ref{tab:cot-vs-constcot} summarizes the key differences between Const-o-T and CoT across representation, purpose, efficiency, and robustness.


\textbf{Modifying the Planner.}
\textcolor{black}{A key distinction of Const-o-T is that it modifies the planner itself, rather than guiding it externally. Prior LLM-based approaches (e.g., CoT, ToT, verifier-based methods) influence action selection through scoring, prompting, or filtering, but still operate over the full action space $A(s)$. In contrast, each constraint $\kappa_t$ explicitly defines a restricted action set $A_{\kappa_t}(s) \subseteq A(s)$, and this restriction is enforced before node expansion. From a search perspective, this mechanism dynamically redefines the branching factor at each step: instead of exploring $|A(s)|$ actions, the planner operates over $|A_{\kappa_t}(s)|$, where $|A_{\kappa_t}(s)| \ll |A(s)|$. As a result, Const-o-T alters the structure of the search tree itself, eliminating infeasible branches before they are explored.}

\textbf{Constraints as Search Controllers.} 
 \textcolor{black}{In Const-o-T, constraints restrict the admissible action set during search, rather than filtering outputs post-hoc. Each constraint $\kappa_t$ induces a reduced action space $\mathcal{A}_{\kappa_t}(s)$, ensuring that only feasible actions are considered during node expansion. This transforms reasoning from descriptive guidance into a control mechanism: constraints act as search controllers that shape exploration in real time, preventing infeasible actions from entering the search tree. In contrast, post-hoc filtering cannot influence exploration and often leads to wasted computation or failed search trajectories. Our contribution is not post-hoc filtering, but defining the admissible action set $\mathcal{A}_{\kappa_t}(s)$ during search. Importantly, the method is robust to moderate constraint errors, as the search process degrades gracefully rather than failing catastrophically when constraints are imperfect.}


\textbf{Beyond Accuracy Improvements.} While improvements in final task accuracy are sometimes modest, Const-o-T consistently reduces search complexity and, in interactive evaluation, substantially improves perceived alignment between user intent and system behavior (See Figure~\ref{fig:user_study}). This pattern is consistent with the role of executable constraints as search controllers: their primary effect is to restrict infeasible reasoning paths and preserve intent during planning.  \textcolor{black}{Furthermore, our approach can be viewed as a budgeted, constraint-guided variant of MCTS, where constraints reduce the effective branching factor of the search, leading to more efficient and focused exploration.}

\vspace{-12pt}
\section{Limitations}
\vspace{-10pt}
Const-o-T depends on accurate constraint extraction; when LLMs misinterpret user input or produce incomplete symbolic constraints, the search may be guided toward suboptimal solutions. Additionally, the framework assumes users can articulate intent in natural language; however, for complex scenarios, users may struggle to express nuanced preferences, potentially leading to constraint extraction failures and human-AI alignment issues.

\vspace{-12pt}
\section{Conclusion}
\vspace{-10pt}
We introduced Constraints-of-Thought (Const-o-T), a framework that transforms unconstrained natural language reasoning into structured, verifiable constraints for guiding language model planning. By representing each reasoning step as an $\langle$intent, constraint$\rangle$ pair, our approach provides actionable control over the search process rather than merely explanatory rationales. Our integration with MCTS demonstrates how constraints can effectively prune infeasible branches and direct exploration toward semantically meaningful actions. Comprehensive evaluation across four domains shows consistent improvements in accuracy, structural alignment, and constraint satisfaction. These findings suggest that structured constraint extraction represents \textcolor{black}{a promising direction for constraint-guided reasoning in LLM-based planning systems.}

\section*{Acknowledgment}
This work was supported by the Office of Naval Research (ONR) under grant number N00014-23-1-2887, by the National Science Foundation under grant number CMMI-2229260, and by a gift from Konica Minolta.


\bibliographystyle{unsrtnat}
\bibliography{custom}







\appendix

\section{Risk Game environment}

Risk is a strategy-based game originally developed as a board game. Our online adaptation faithfully preserves its core rules, challenging players in diplomacy, territorial conquest, and conflict resolution. The objective is to achieve world domination across a custom-designed map composed of 6 continents and 21 territories. Players take turns deploying troops and attempting to capture territories from their opponents, with combat outcomes determined by dice rolls.

\begin{figure}[H]
  \centering
  \includegraphics[width=\linewidth]{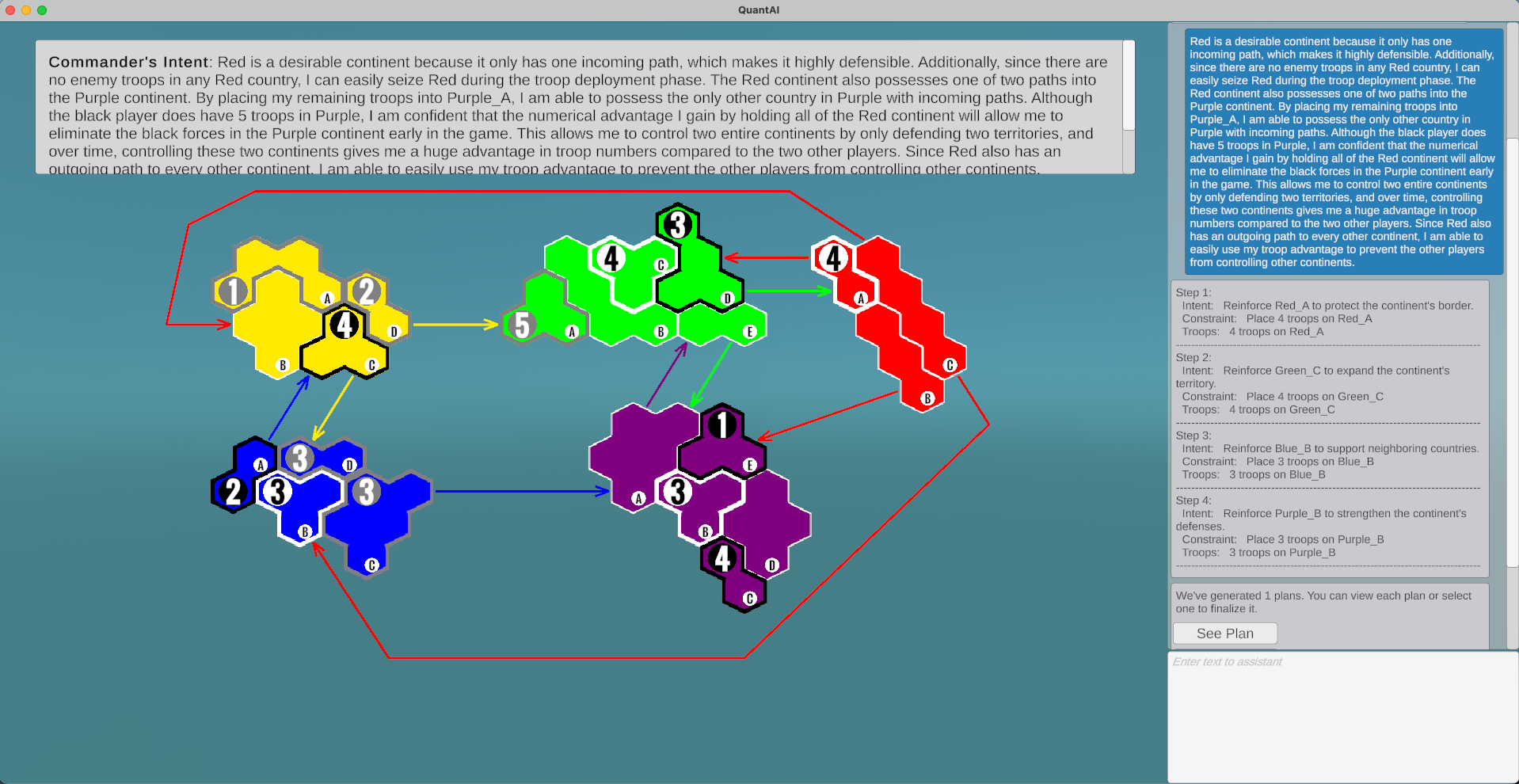}
    \caption{Risk game – Turn 0, showing the initial troop placement by the player.}

  \label{fig:schematicRisk_1}
\end{figure}

\begin{figure}[H]
  \centering
  \includegraphics[width=\linewidth]{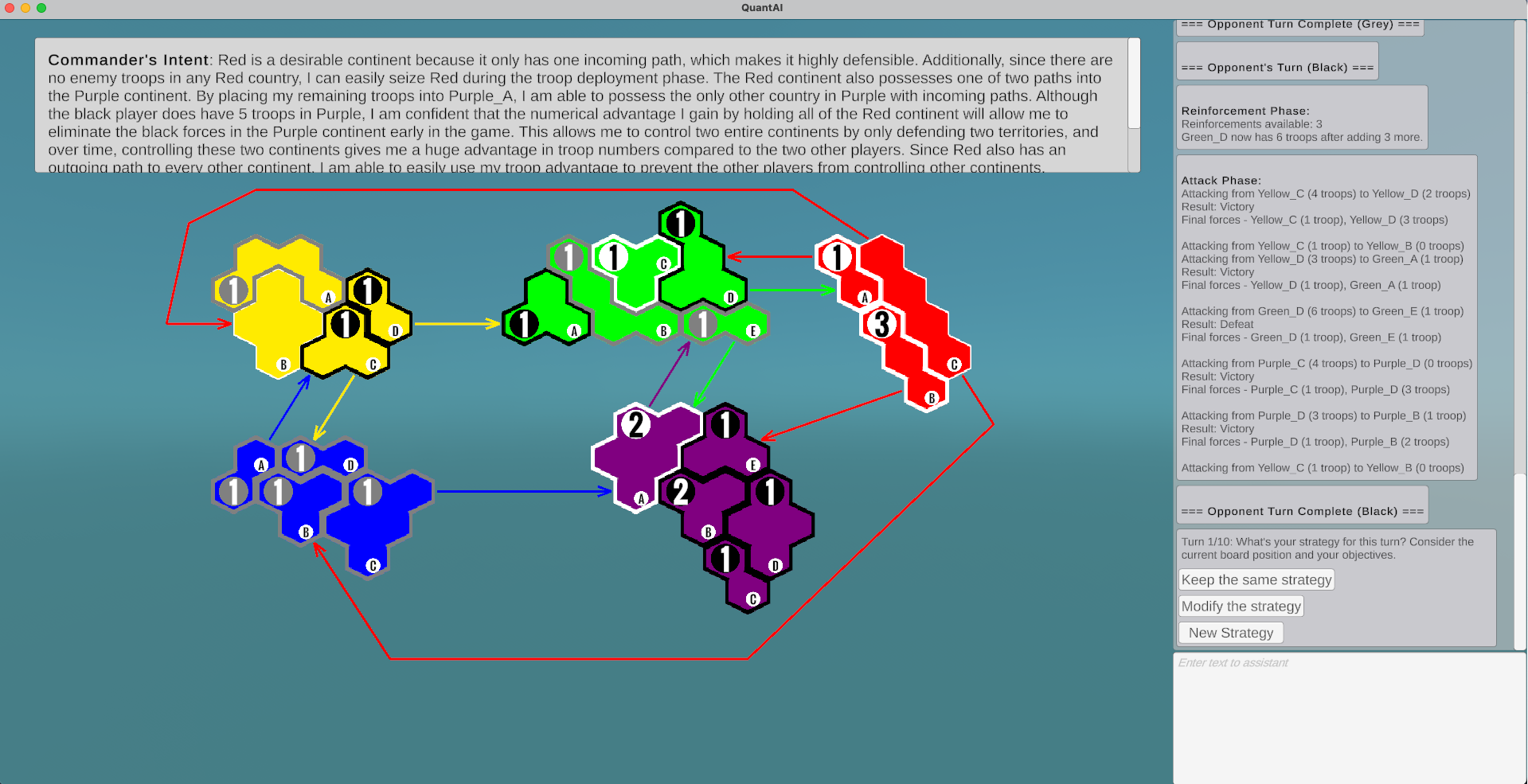}
    \caption{Risk game: Post–Turn 0, as the Black and Gray players make their moves.}

  \label{fig:schematicRisk_2}
\end{figure}

In our setup, the participant plays as the White player, while two heuristic agents control the Black and Grey factions. The game begins with the participant allocating troops to preferred regions by interacting through an AI-powered chat interface. An AI planner then generates a proposed plan that aligns with the participant’s stated intent. The participant can either accept this plan or provide feedback to update it, as illustrated in Fig.~\ref{fig:schematicRisk_1}-\ref{fig:schematicRisk_2}.

Once the participant lays down their troops, the opponents make their moves. Once the turn circles back to the participant, as seen in Fig.~\ref{fig:schematicRisk_2}, they can either use the same strategy as before or give a new one to the planner.

\begin{figure}[H]
  \centering
  \includegraphics[width=\linewidth]{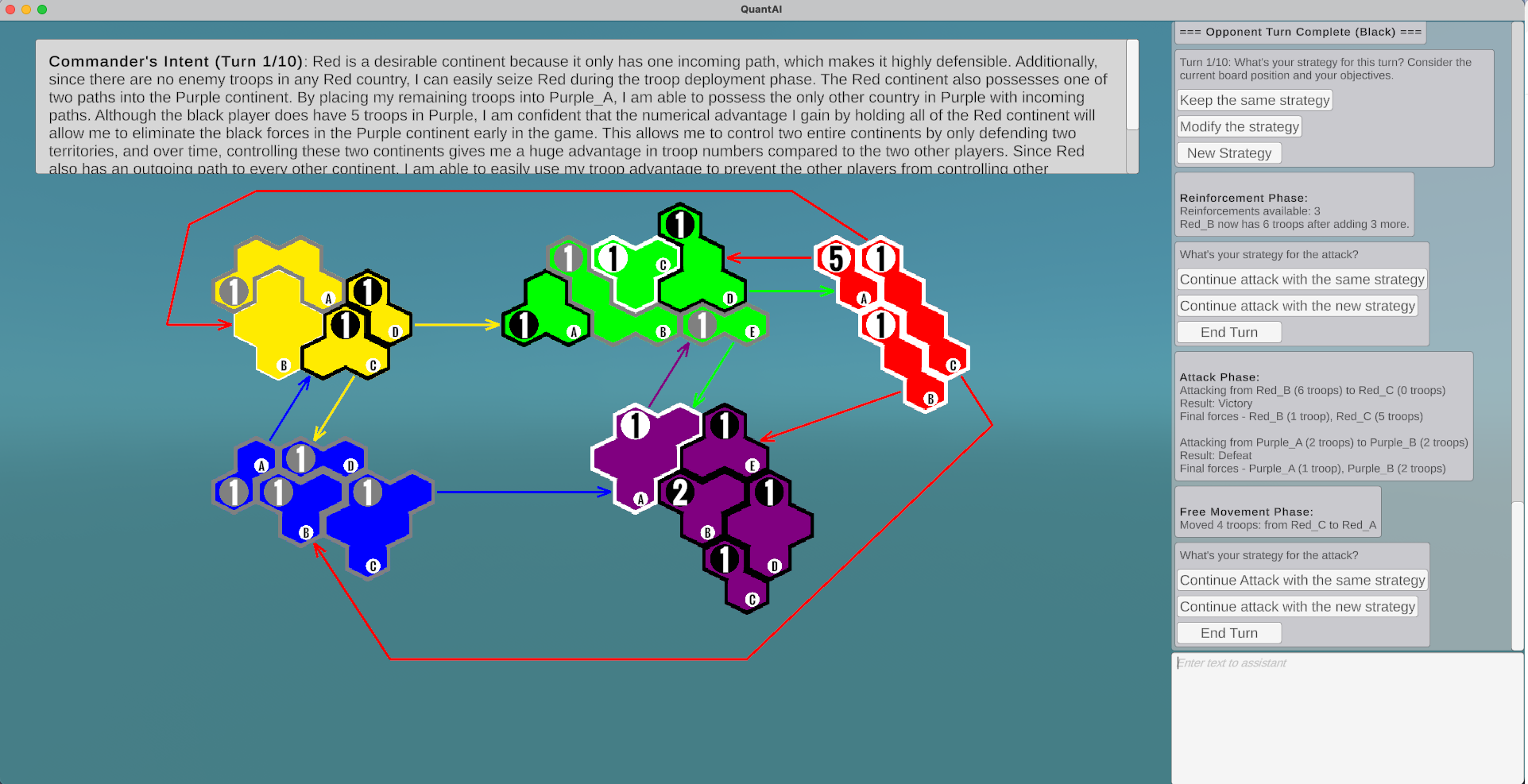}
    \caption{Risk game: Turn 1, where the player proceeds through the reinforcement, attack, and freemove phases.}

  \label{fig:schematic Risk_3.png}
\end{figure}

Next, the participant can choose to Reinforce, Attack, or Freemove (as shown in Fig.~\ref{fig:schematic Risk_3.png}). At the start of each turn, the player receives reinforcement armies proportional to the number of territories they control, with additional bonus armies granted for holding entire continents. These reinforcements can be used to strengthen key strongholds.

Players may attack adjacent or connected opponent territories—those linked by unidirectional arrows. The outcomes of attacks are determined by dice rolls, with each roll resulting in the loss of a certain number of troops by either the attacker or defender. The more troops committed to an attack, the higher the chances of success. A battle continues until the attacker chooses to stop, runs out of armies to attack with, or successfully eliminates the last defending unit—at which point they take over the territory by moving armies into it.

At the end of the turn, the player may perform a Freemove, redistributing armies between their own connected territories. This cycle repeats until one player achieves world domination.


\begin{algorithm}
\caption{Constraints-of-Thought (Const-o-T) Extraction from user input}
\label{alg:constcot-extract}
\begin{algorithmic}[1]
\Require User description $D$; map state $M$; prompt template $\mathcal{T}$
\Ensure Sequence $\mathcal{C} = [c_i]_{i=1}^K$, where $c_i = \langle i_i, \kappa_i \rangle$; total count $K$
\Statex

\Function{ConstCoT\_Extract}{$D, M, \mathcal{T}$}
    \State $\widehat{\mathcal{C}} \gets LM(D, M, \mathcal{T})$
    \State $\mathcal{C} \gets [\,]$
    \ForAll{$c \in \widehat{\mathcal{C}}$}
        \If{\Call{Validate}{$c$}}
            \State $\mathcal{C} \gets \mathcal{C} \cup [c]$
        \EndIf
    \EndFor
    \State $K \gets |\mathcal{C}|$
    \State \Return $(\mathcal{C}, K)$
\EndFunction
\Statex

\Function{Validate}{$c$}
    \State \textbf{require} fields present: step\_id, $i_t$ (natural-language intent), $\kappa_t$ (constraint predicate)
    \State \textbf{require} $i_t$ length within bounds; $\kappa_t$ matches grammar $\mathcal{G}$ or schema $\mathcal{S}$
    \State \textbf{require} $\kappa_t$ is feasible under $X'$ if available (e.g., legal action, resource limits)
    \State \Return true if all checks pass; else false
\EndFunction
\Statex
\end{algorithmic}
\end{algorithm}

\section{Constraints-of-Thought (Const-o-T)}
\label{App_const}
The key idea behind Const-o-T is to transform natural language strategies into a sequence of structured $\langle \text{intent}, \text{constraint} \rangle$ pairs, where each constraint is an executable predicate that directly restricts the planner’s feasible action space. Formally, each constraint $\kappa_t$ defines a predicate over actions that induces a restricted action set $A_{\kappa_t}(s) \subseteq A(s)$, and this restriction is enforced \emph{before node expansion} during search. This transforms intermediate reasoning from descriptive rationales into \emph{search controllers} that modify the planner itself.

These extracted constraints serve two roles. First, they align planning with the user’s high-level intent by restricting the admissible actions to those consistent with the intended strategy. Second, they ensure feasibility by enforcing domain-specific requirements (e.g., game rules or geometric validity) through symbolic validation. 

Algorithm~\ref{alg:constcot-extract} illustrates this process. In line 2, the language model maps the user input to a sequence of $\langle \text{intent}, \text{constraint} \rangle$ pairs. From lines 4 to 10, each candidate constraint is validated to ensure structural correctness and feasibility. The resulting constraint set defines the admissible action space used during downstream reasoning or planning, ensuring that only constraint-consistent actions are considered throughout the search process.



\begin{algorithm*}[H]
\caption{Constraints-of-Thought (Const-o-T) Extraction from user input}
\label{alg:constcot-extract}
\begin{algorithmic}[1]
\Require User description $D$; map state $M$; prompt template $\mathcal{T}$
\Ensure Sequence $\mathcal{C} = \big[(\text{intent}_i, \text{constraint}_i)\big]_{i=1}^K$; total count $K$
\Statex

\Function{ConstCoT\_Extract}{$D, M, \mathcal{T}$}
    \State $\widehat{\mathcal{C}} \gets LM(D, M, \mathcal{T})$
    \State $\mathcal{C} \gets [\,]$
    \ForAll{$c \in \widehat{\mathcal{C}}$}
        \If{\Call{Validate}{$c$}}
            \State $\mathcal{C} \gets \mathcal{C} \cup [c]$
        \EndIf
    \EndFor
    \State $K \gets |\mathcal{C}|$
    \State \Return $(\mathcal{C}, K)$
\EndFunction
\Statex

\Function{Validate}{$c$}
    \State \textbf{require} fields present: step\_id, intent (natural language), constraint (formal/actionable)
    \State \textbf{require} intent length within bounds; constraint matches grammar $\mathcal{G}$ or schema $\mathcal{S}$
    \State \textbf{require} constraint feasibility under $X'$ if available (e.g., legal action, resource limits)
    \State \Return true if all checks pass; else false
\EndFunction
\Statex

\end{algorithmic}
\end{algorithm*}

\section{User Study}
\label{User_Study_app}
The user study was conducted with 18 participants (5 female and 13 male). All participants over the age of 18 were welcome but the average age of the participants was 21.7. Participants had varying levels of experience with AI systems, gaming, strategy-based games and Risk. Trust scale items were adapted from validated measures of trust in automation~\citep{jian2000foundations,kizilcec2016much} and technology acceptance~\citep{davis1989perceived}, with wording contextualized to the Risk domain (e.g., “I trust the system’s troop placement/plan suggestions”). \textcolor{black}{Usability scale items were also contextualized to the Risk domain (e.g., ``I could easily predict how changes in my strategy description would affect the system’s decisions''), following prior work on user understanding of AI behavior~\citep{liao2020questioning}.} The reliability of each metric was calculated using Cronbach's $\alpha$ and reported in Table \ref{calpha}. Normality and heteroskedasticity assumptions were checked using a Shapiro-Wilk test and a Levene's test, respectively. Transparency and usability metrics passed these tests. The results of the ANOVA and post-hoc Tukey's HSD test for transparency and usability and the Kruskal-Wallis test and post-hoc Dunn's test for trust and alignment are reported in Table \ref{anova} and Table \ref{dunn} respectively. 

\begin{table}[htbp]
\centering
\caption{Reliability analysis by metric and mode. $\alpha$ $>$ 0.9 has excellent reliability, $\alpha$ $>$ 0.8 has good reliability and $\alpha$ $>$ 0.7 has acceptable reliability}

\small
\begin{tabular}{llcc}
\hline
\textbf{Metric} & \textbf{Mode} & \textbf{$\alpha$} & \textbf{Reliability} \\
\hline
\multirow{3}{*}{Transparency} 
 & Mode 1 & 0.95 & Excellent \\
 & Mode 2 & 0.93 & Excellent \\
 & Mode 3 & 0.95 & Excellent \\
\hline
\multirow{3}{*}{Usability} 
 & Mode 1 & 0.81 & Good \\
 & Mode 2 & 0.76 & Acceptable \\
 & Mode 3 & 0.78 & Acceptable \\
\hline
\multirow{3}{*}{Trust} 
 & Mode 1 & 0.79 & Acceptable \\
 & Mode 2 & 0.93 & Excellent \\
 & Mode 3 & 0.90 & Excellent \\
\hline
\end{tabular}
\label{calpha}
\end{table}

\begin{table*}[h]
\centering
\caption{ANOVA and post-hoc results for transparency and usability}

\small
\begin{tabular}{lcc}
\hline
 & Transparency & Usability \\
\hline
APA & $F(2, 51) = 7.78, \; p < 0.01 \;$ & $F(2, 51) = 10.23, \; p < 0.001 \;$ \\
Shapiro $p$ & 0.427 & 0.971 \\
Levene $p$ & 0.996 & 0.843 \\
Eta$^{2}$ & 0.234 & 0.286 \\
Power & 0.951 & 0.987 \\
\hline \\
\multicolumn{3}{l}{\textbf{Tukey HSD pairwise comparisons}} \\
\hline
Comparison & Transparency (mean diff, $p$) & Usability (mean diff, $p$) \\
\hline
Mode 1 vs Mode 2 & -1.21, $p=0.001$ ** & -1.83, $p=0.0001$ *** \\
Mode 1 vs Mode 3 & -0.84, $p=0.026$ * & -1.07, $p=0.030$ * \\
Mode 2 vs Mode 3 & +0.36, $p=0.481$ n.s. & +0.76, $p=0.156$ n.s. \\
\hline \\
\multicolumn{3}{l}{\textbf{Cohen's $d$ (pairwise effect sizes; d $>$ 0.8 signifies large effect size)}} \\
\hline 
Comparison & Transparency & Usability  \\
\hline 
Mode 1 vs Mode 2 & 1.29 & 1.52 \\
Mode 1 vs Mode 3 & 0.90 & 0.89 \\
Mode 2 vs Mode 3 & -0.39 & -0.61 \\
\hline
\end{tabular}
\label{anova}
\end{table*}

\begin{table*}[h]
\centering
\caption{Kruskal–Wallis tests, Dunn post-hoc comparisons, and Cohen’s d effect sizes for Trust and Alignment.}

\small
\begin{tabular}{lcc}
\hline
 & Trust & Alignment \\
\hline
APA  & $H(2) = 19.80, \; p < 0.001 \; $ & $H(2) = 23.75, \; p < 0.001 \;$ \\
\hline \\
\multicolumn{3}{l}{\textbf{Dunn pairwise comparisons (adjusted $p$)}} \\
\hline
Comparison & Trust & Alignment \\
\hline
Mode 1 vs Mode 2 & $p = 0.0001$ *** & $p < 0.00001$ *** \\
Mode 1 vs Mode 3 & $p = 0.0014$ ** & $p = 0.0011$ ** \\
Mode 2 vs Mode 3 & $p = 1.000$ n.s. & $p = 0.814$ n.s. \\
\hline \\
\multicolumn{3}{l}{\textbf{Cohen's $d$ (pairwise effect sizes; d $>$ 0.8 signifies large effect size)}} \\
\hline
Comparison & Trust & Alignment \\
\hline
Mode 1 vs Mode 2 & 1.76 & 2.19 \\
Mode 1 vs Mode 3 & 1.51 & 1.67 \\
Mode 2 vs Mode 3 & -0.26 & -0.43 \\
\hline
\end{tabular}
\label{dunn}
\end{table*}

\textcolor{black}{We conducted a one-way ANOVA for transparency and usability and a Kruskal–Wallis test for trust and alignment. Significant differences across interaction modes were observed for transparency ($F(2,51)=7.78$, $p<0.01$), usability ($F(2,51)=10.23$, $p<0.001$), trust ($H(2)=19.80$, $p<0.001$), and alignment ($H(2)=23.75$, $p<0.001$), as shown in Fig.~\ref{fig:user_study}. Post-hoc pairwise comparisons (Tukey HSD and Dunn tests) indicate that the aligned condition significantly outperformed both agnostic and adversarial conditions across all metrics.
These results indicate that constraint-guided planning produces actions that users perceive as significantly more aligned with their stated strategies than systems that either ignore or oppose user intent.}

\section{Additional Analysis}

\begin{figure*}[h]
\centering
\scriptsize
\begin{tcolorbox}[
    title=Examples of Failure Modes from Risk game,
    colback=gray!5,
    colframe=black,
    colbacktitle=gray!20,
    coltitle=black,
    sharp corners,
    width=.99\linewidth,
]
\textbf{1. Wrong territory selected by the extracted constraint}
\\
\textbf{Example 1: }
\\
I need to defend my troops in major areas like Purple and Yellow.  By placing my troops in these areas I can move them freely to attack my opponent and win the battles.  I also need to have more troops on the board to help gain more territory.  I am want to fortify my troops so i moved more troops to boarders.  This way I can move my troops and attack freely.  this will help attack and gain ground.  

GT: [('Green\_C', 4), ('Blue\_B', 3), ('Green\_B', 4), ('Yellow\_B', 1), ('Purple\_D', 2)]

Pred: [('Purple\_A', 8), ('Yellow\_B', 6)]
\\
\textbf{Example 2: }
\\
I'm looking to kick the other player off of purple e and then make my way over to red b to attack and take that continent. I don't see a threat from blue yet but also want to be able to defend and attack from green if they make a move on me. 

GT: [('Purple\_B', 5), ('Purple\_C', 5), ('Purple\_D', 4)]

Pred: [('Purple\_A', 6), ('Red\_B', 5), ('Green\_B', 3)]

\textbf{2. Constraint too restrictive, eliminating a good move }
\\
\textbf{Example 1: }
\\
The yellow contenint is my get away, I can get to any country by the yellow. Thats why I start off on yellow because its prepared to take out any other contenient. The red contienent is what Id end om bexause the red is the ending of the game 
GT: [('Yellow\_D', 3), ('Green\_B', 3), ('Blue\_B', 3), ('Purple\_B', 3), ('Blue\_D', 2)]

Pred: [('Yellow\_D', 14)]
\\
\textbf{Example 2: }
\\
Red I feel is the central part of the map. I almost even believe that is why the map shows it so far out to trick you to think it is not. I think Red would be the place to go, to make sure you can get to everyone quickly and start to take over each continent one at a time. 
\\
GT: [('Blue\_A', 1), ('Purple\_B', 1), ('Yellow\_A', 2), ('Red\_B', 2), ('Blue\_B', 2), ('Blue\_C', 4), ('Blue\_D', 2)]

Pred: [('Red\_B', 7), ('Red\_C', 7)]

\textbf{3. Vague input leading to incomplete structure }
\\
\textbf{Example 1: }
\\
As I looked at the board I wanted to spread my troops to three continents. On those continents I was just going to occupy one country each. Doing it that way would give me a chance to be able to attacked continent purple from two different continents. I would also be able to attack continent green from a couple different continents. Giving me the opportunity to be able to make these attacks and perhaps win them. From there I could march on and keep battling troops and conquering more countries. 
\\
GT: [('Purple\_A', 4), ('Red\_B', 5), ('Blue\_C', 5)]

Pred: [('Blue\_A', 5), ('Red\_C', 4), ('Purple\_A', 5)]

\textbf{Example 2: }
\\
I am planning a focused strategy, by limiting deployment of troops in terms of number of countries and continents. I am avoiding battle on territories that are already heavily occupied, since I lack confidence and don't want to immediately lose a lot of troops. I want to establish a stronghold in at least one country in at least 2 territories. That way, I can act as a spoiler to keep other players from taking over the whole board. I chose countries that offered some mobility because of the presence of an arrow. Once I establish a foothold in my initial continent choices, I will move on toward world domination. 

GT: [('Green\_E', 5), ('Blue\_A', 5), ('Yellow\_D', 4)]

Pred: [('Yellow\_D', 7), ('Green\_D', 7)]

\end{tcolorbox}
\caption{Failure Modes}
\label{fig:Failure_Modes}
\end{figure*}


\textbf{Constraint Quality.} To evaluate the quality of the extracted constraints, we compare the predicted constraint set against the ground-truth troop placements in the Risk dataset. We compute standard classification metrics over territory-level constraint predictions. Our method achieves 0.82 precision, and 0.81 recall, indicating that the extracted constraints are both highly precise and largely complete. Manual inspection of 100 randomly sampled examples indicates that most errors arise from ambiguous strategy descriptions or underspecified user goals.

\textbf{Statistical Analysis.} We conducted one-way ANOVA tests to evaluate how different experimental settings affect performance for GPT-4.1. For search methods (MCTS, MCTS + CoT, and MCTS + Const-o-T), we observe a significant effect on EMD ($F(2,3156)=75.92$, $p < 0.001$, $\eta^2 = 0.046$, $f = 0.219$), \textcolor{black}{indicating that reasoning strategy significantly affects placement quality.}. Further, prompting approaches also show differences ($F(7,8377)=15.94$, $p < 0.001$, $\eta^2 = 0.013$, $f = 0.115$).

\begin{table}[t]
\centering
\caption{Ablation: intent and constraint components}

\setlength{\tabcolsep}{4pt}
\renewcommand{\arraystretch}{0.9}
\begin{tabular}{lc}
\toprule
Variant & EMD $\downarrow$ \\
\midrule
w/o Intent & 0.53 \\
w/o Constraint & 0.52 \\
w/o Verification & 0.53 \\
w/o Intent \& Constraint & 0.79 \\
\midrule
\textbf{Ours} & \textbf{0.51} \\
\bottomrule
\end{tabular}
\label{ablation}
\end{table}

\begin{figure}[h]
  \centering
  \includegraphics[width=.5\linewidth]{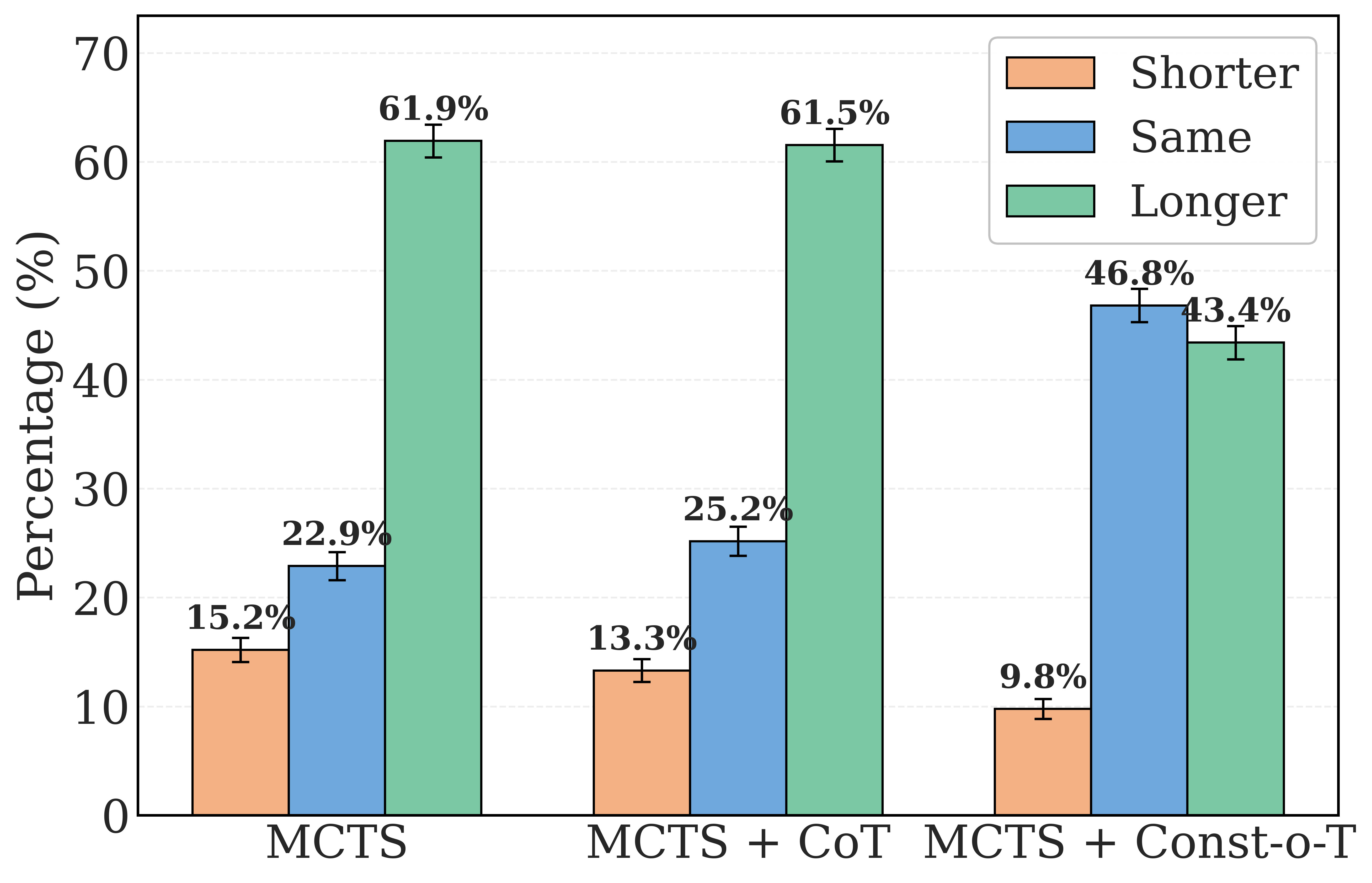}
  \caption{Distribution of plan lengths relative to ground truth for GPT-4.1.}
  \label{fig:plan_length_distribution}
  \vspace{-8pt}
\end{figure}

\textbf{Error Analysis.} \textcolor{black}{Figure~\ref{fig:plan_length_distribution} shows that classical MCTS and MCTS with CoT often over-generate plans, whereas MCTS with Const-o-T better aligns plan length with the ground truth. This behavior arises because constraints restrict the admissible action set during search, which reduces the branching factor by preventing unnecessary expansions. As a result, Const-o-T operates as a budgeted, constraint-guided variant of MCTS that focuses exploration on feasible and relevant actions, reducing over-generation.}

\textbf{Ablation Study.} We conducted an ablation study on 200 GPT-4.1 examples to examine the contributions of intent extraction, constraint enforcement, and verification. Removing intent increases EMD from 0.51 to 0.53, while removing constraints increases it to 0.52. Disabling the verification step yields a similar EMD of 0.53. In contrast, removing both intent and constraint substantially degrades performance, increasing EMD to 0.79. These results indicate that each component contributes to improving alignment with ground-truth placements, while the combination of intent and constraint is critical for maintaining stable search behavior. The full model achieves the best performance (EMD = 0.51; Table~\ref{ablation}). While verification contributes to performance, the primary gains arise from restricting the action space during search. Verification acts as a safeguard ensuring constraint satisfaction rather than the primary driver of improvements.

\section{Fitness Function}
\label{Fitness_function}

For the game of \textit{Risk}, during the troop deployment phase, we evaluate a candidate deployment state $v$ using a fitness function that balances goal satisfaction against constraint violations. The fitness function used to evaluate a candidate deployment state, $v$, is defined in Eq.~\ref{eq:fitness_function}.
\begin{equation}
V(v) \;=\; \sum_{i=1}^6 w_i \, g_i(v) \;-\; 
\lambda \sum_{m=1}^9 c_m(v),
\label{eq:fitness_function}
\end{equation}
In this equation, $g_i(v)$ denotes normalized goal scores that measure how well the configuration satisfies strategic objectives. The coefficients $w_i$ denote non-negative weights that encode the relative importance of each goal. The functions $c_m(v)$ denote binary indicators of constraint violations. The penalty coefficient, $\lambda$, is set sufficiently large to ensure that any violation dominates the weighted sum of goals, thereby prioritizing feasible deployments over infeasible ones.

\paragraph{Goals.}
The goal functions $g_i(v)$ measure how well the player’s configuration $v$ satisfies strategic objectives:
\begingroup
\small
\setlength{\jot}{4pt} 
\begin{align}
g_1(v) &= \frac{\#\{\text{occupied territories adjacent to enemy}\}}{\#\{\text{occupiable territories adjacent to enemy}\}}, \\[2pt]
g_2(v) &= \frac{\#\{\text{countries controlled}\}}{\#\{\text{countries occupiable}\}}, \\[2pt]
g_3(v) &= 1 - \frac{1}{|\mathcal{T}(v)|\big(|\mathcal{T}(v)|-1\big)}
           \sum_{\mathclap{\substack{u,w \in \mathcal{T}(v)\\ u \neq w}}}
           \frac{d(u,w)}{d_{\max}}, \\[2pt]
g_4(v) &= \frac{\#\{\text{troops adjacent to enemy}\}}{\#\{\text{total troops}\}}, \\[2pt]
g_5(v) &= \frac{\#\{\text{border troops on controlled continents}\}}{\#\{\text{troops on controlled continents}\}}, \\[2pt]
g_6(v) &= 1 - \frac{\#\{\text{unique enemy players adjacent}\}}{\#\{\text{maximum enemies}\}}.
\end{align}
\endgroup

\noindent\textit{Descriptions.}
\(g_1\): surround enemy territories; 
\(g_2\): maximize territorial control;
\(g_3\): minimize average pairwise troop distance (uses graph distance \(d(\cdot,\cdot)\), normalized by \(d_{\max}\));
\(g_4\): maximize battles throughout the game;
\(g_5\): fortify borders of continents you control;
\(g_6\): limit exposure to many enemies.

\paragraph{Constraints.}
Constraints are formulated as binary functions that equal $1$ when violated and $0$ otherwise:
\begin{equation}
\begin{aligned}
c_1(v) &= \mathbb{1}[\text{no troop on required continent}] \\
c_2(v) &= \mathbb{1}[\text{troop placed on forbidden continent}] \\
c_3(v) &= \mathbb{1}[\text{cannot reach continent in one move}] \\
c_4(v) &= \mathbb{1}[\text{border of continent not defended}] \\
c_5(v) &= \mathbb{1}[\text{insufficient troops to defend continent}] \\
c_6(v) &= \mathbb{1}[\text{fewer than required countries}] \\
c_7(v) &= \mathbb{1}[\text{troops on fewer than required continents}] \\
c_8(v) &= \mathbb{1}[\text{fewer than required troops per country}] \\
c_9(v) &= \mathbb{1}[\text{troops on more than allowed continents}]
\end{aligned}
\end{equation}

\noindent
Together, these definitions create a fitness landscape in which higher scores correspond to strategically advantageous and constraint-compliant configurations. The weighting scheme allows tailoring the optimization toward different strategic preferences while ensuring that hard constraints remain non-negotiable.

\section{Comparison of Chain-of-Thought (CoT) and Constraints-of-Thought (Const-o-T)}
CoT relies on unconstrained natural language reasoning that helps describe the thought process but lacks verifiability and control over the search process. \textcolor{black}{CoT} is best suited for single-step QA or explanation tasks, but often leads to large search spaces and hallucinations.

In contrast, Const-o-T converts strategies into structured symbolic constraints, which can be formally verified and used to guide or prune search (e.g., in MCTS). This results in more efficient, robust, and goal-aligned planning, making it especially effective for multi-step tasks like strategy games and CAD design.

\begin{table}[h]
\centering
\caption{Comparison of Chain-of-Thought (CoT) and Constraints-of-Thought (Const-o-T).}

\scriptsize
\renewcommand{\arraystretch}{1.15}
\resizebox{\columnwidth}{!}{%
\begin{tabular}{p{2.2cm}p{4.2cm}p{4.5cm}}
\toprule
\textbf{Aspect} & \textbf{Chain-of-Thought (CoT)} & \textbf{Constraints-of-Thought (Const-o-T)} \\
\midrule
Representation & unconstrained natural reasoning trace & Structured symbolic constraints (equations, rules) \\ \midrule
Purpose & Describes reasoning steps & Prescribes feasible solution space \\ \midrule
Search Interaction & Linear expansion; may increase branching & Guides/prunes expansions; controls horizon and branching factor \\ \midrule
Verification & No formal mechanism; correctness by final output & Constraints checkable via solvers/game rules \\ \midrule
Efficiency & Larger search space; redundant paths possible & Compressed space; faster convergence \\ \midrule
Robustness & Prone to hallucinations and drift & Enforces consistency; fewer invalid solutions \\ \midrule
Use Case Fit & Best for single-step QA/explanation & Suited for multi-step planning, strategy, CAD/game tasks \\ \midrule
Novelty & Thoughts as \emph{rationales} & Thoughts as \emph{controllers} (constraints for symbolic search) \\
\bottomrule
\end{tabular}}
\label{tab:cot-vs-constcot}
\end{table}

\section{Commander's Intent Dataset Example}
\label{CI_examples}
\begin{figure*}[h]
\centering

\begin{tcolorbox}[title=Language strategy., colback=gray!5, colframe=black, 
  colbacktitle=gray!20, coltitle=black, sharp corners, width=.99\linewidth]
This one was difficult but I determined that controlling red would allow me the greatest chance of success as it provides a good base for defense while also granting large movement opportunities to attack most of the board. Green will be the biggest challenge as they have the most number of troops with 20 that they can move around, so the strategy would be to gain control of red as quickly as possible before moving in on yellow while leaving the borders of red as strong as possible. Gaining control of yellow and red would allow me to keep green on the defense from opposite sides, but give me access to work control of blue as green will most likely make moves to gain purple as a stronghold. The trick will be playing a long game to maintain control of the split continents while green will ultimately control the center of the board until I can attempt to force them down into the purple region and attack from several sides.
\end{tcolorbox}

\vspace{0.5em} 

\includegraphics[width=\linewidth]{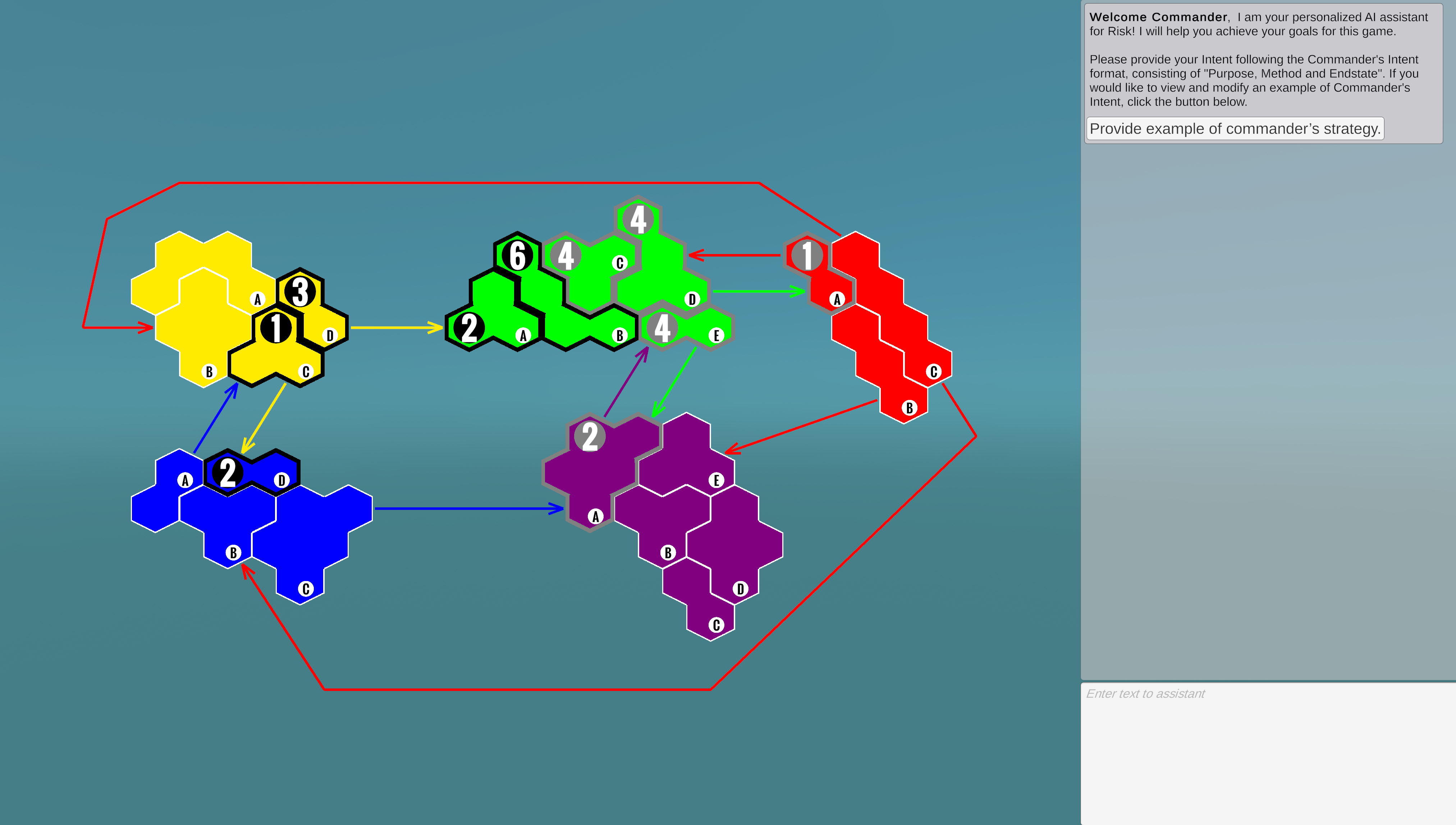}

\caption{An example from the Commander’s Intent dataset. 
Top: a natural language strategy description. 
Bottom: the corresponding map scenario. 
The ground truth troop placements are Country = Red\_B with 7 troops 
and Country = Red\_C with 7 troops.}
\label{fig:combined_strategy_map}
\end{figure*}

This example from the Commander's Intent dataset highlights the complex strategic reasoning required in a Risk board game scenario (See Figure \ref{fig:combined_strategy_map}). The natural language strategy demonstrates multi-step task planning, where the player first analyzes the current board state and then formulates a plan. The accompanying map depicts the actual game state, with the ground truth showing optimal troop placements of seven units each in Red\_B and Red\_C. These placements align with the strategic focus on securing the Red continent, as outlined in the textual strategy.

\section{Constraint-Optimization Baseline for Troop Placement}
\label{appendix:co_baseline}
We also implemented a constraint optimization based baseline. It is a two-step strategy. In the first step, the commander’s intent is provided to a large language model (we use ChatGPT 4.0 mini) to identify the set of active constraints. These constraints are then enforced using Google’s OR-Tools library \citep{ortools} to trim the search space and obtain a feasible set of troop deployments. If the constraints predicted by the LLM yield no feasible solutions, we retain the largest subset of constraints that produces a feasible solution.  

In the second step, all deployments in the feasible set are evaluated using only the goal component of the fitness function, as defined in Eq.~\ref{eq:fitness_function_goals}.
\begin{equation}
V(v) \;=\; \sum_{i=1}^6 w_i \, g_i(v),
\label{eq:fitness_function_goals}
\end{equation}

In this equation, $g_i(v)$ denotes goal scores that measure how well the deployment state $v$ satisfies strategic objectives. The coefficients $w_i$ denote weights provided by the LLM based on the commander’s intent expressed in natural language, reflecting the relative importance of each objective. The deployment with the highest value of $V(v)$ is selected as the baseline solution.

\begin{table}[t]
\centering
\caption{Results of the constraint-optimization baseline.}
\small
\begin{tabular}{lcc}
\hline
Method & F1 (\%) & Accuracy (\%) \\
\hline
\makecell[l]{Constraint-Optimization\\Baseline} & 57.0 & 62.8 \\
\hline
\end{tabular}
\label{tab:cobaseline}
\end{table}

The performance of the constraint-optimization baseline is shown in Table~\ref{tab:cobaseline}. Even though it leverages constraint-based optimization, its performance compared to \textcolor{black}{human-provided troop-placement ground truth} is significantly worse than our method (See Section~4). In addition, it is not very flexible: each new domain requires hand-crafting a constraint model and solver formulation, whereas our approach transfers more easily and remains flexible across a wide range of applications.

\section{Examples from CAD code generation from CADCodePrompt dataset}
\label{CAD_examples}
\begin{figure*}[h]
    \centering
     \begin{subfigure}[t]{1\linewidth}
        \centering
        \begin{tcolorbox}[colback=gray!5, colframe=black, colbacktitle=gray!20, coltitle=black, sharp corners, width=\linewidth,]
            Write Python code using CADQuery to create a triangular 3D object. First, draw a sketch of an equilateral triangle, pointing downwards. Next, cutout a semicircle from the bottom corner of the triangle. The diameter of this semicircular cutout should be approximately 2/3rd of the length of each side of the triangle. Finally, extrude this sketch to create a 3D object.
        \end{tcolorbox}
        \caption{The natural language descriptions of the 3D object.}
        \label{fig:prompt2}
    \end{subfigure}
    \begin{subfigure}[t]{0.70\linewidth}
        \centering
        \includegraphics[width=\linewidth]{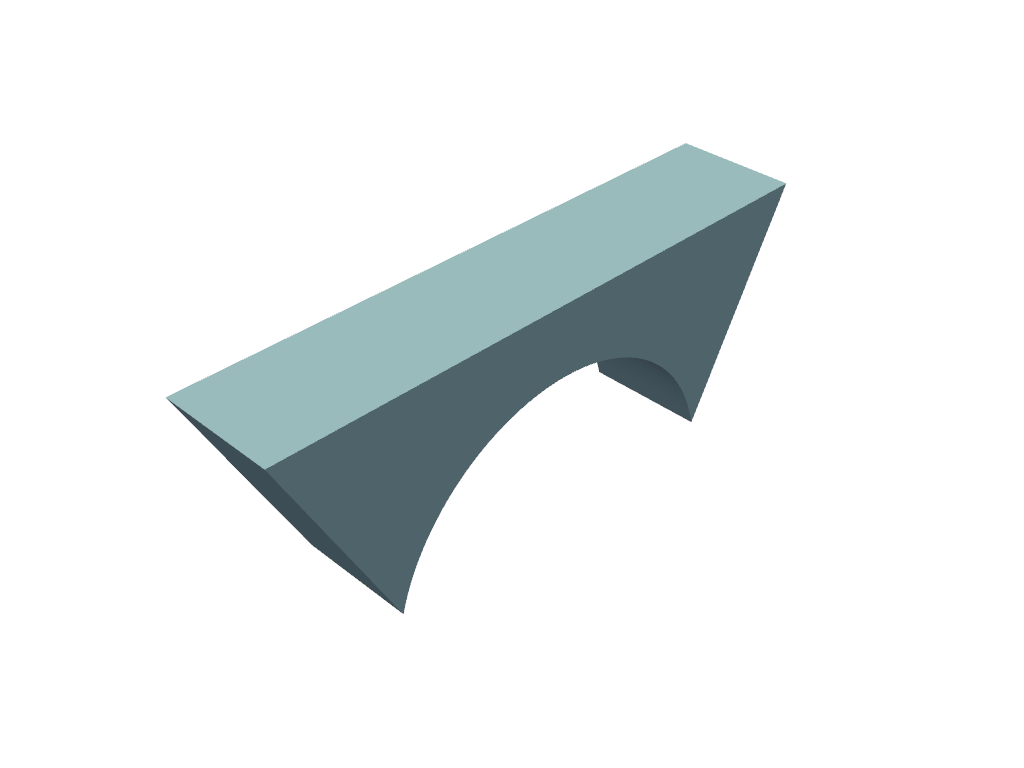}
        \caption{CAD object}
        \label{fig:image_verification_1}
    \end{subfigure}
    \hfill
    \begin{subfigure}[t]{0.70\linewidth}
        \centering
        \includegraphics[width=\linewidth]{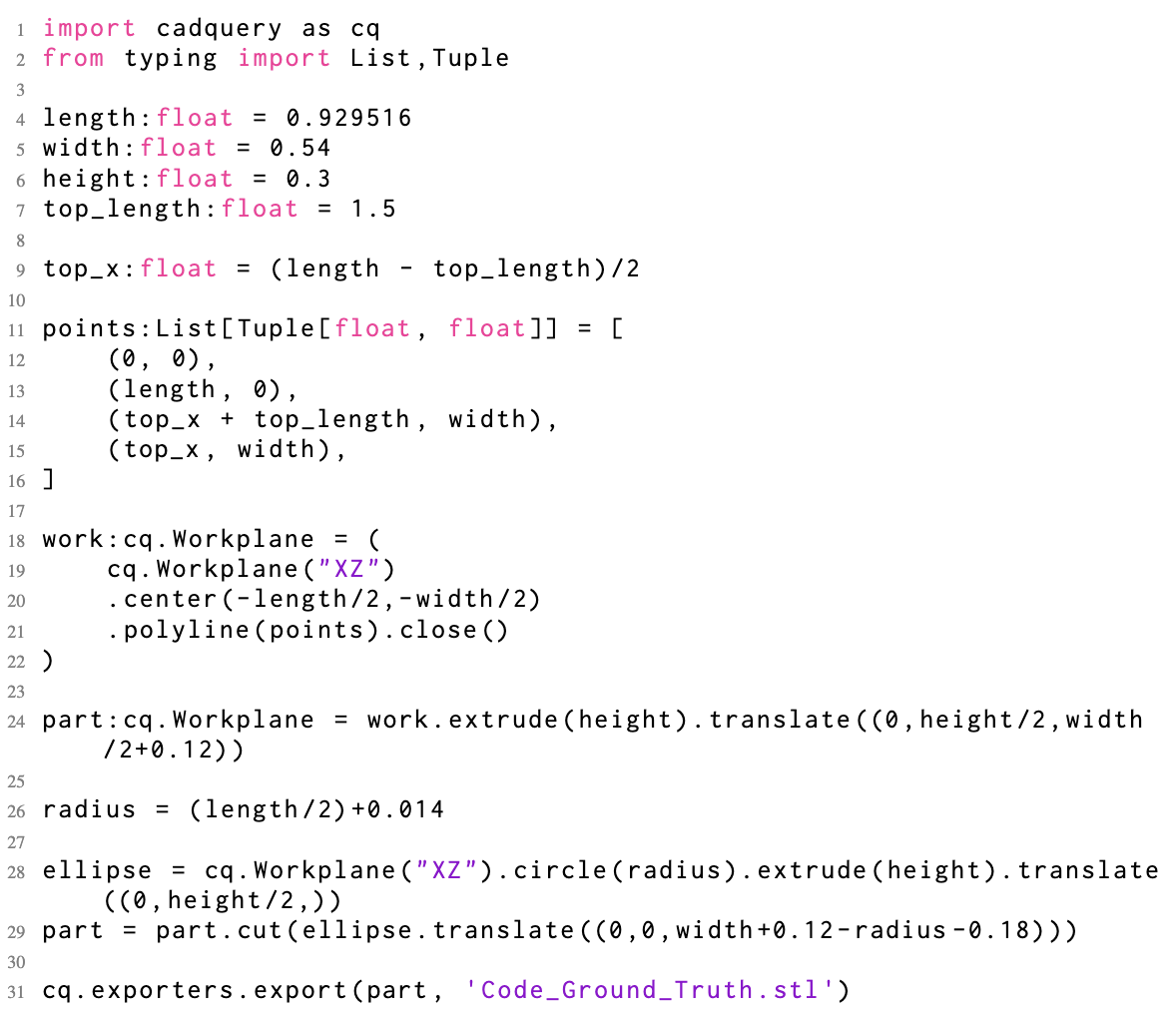}
        \caption{Python Code}
        \label{fig:code_00521217}
    \end{subfigure}
    \hfill
   
    \caption{An example from the \textit{CADPrompt} dataset, showing (a) the prompt, (b) the corresponding CAD object, and (c) the human-annotated Python code used to generate the CAD object.}
    \label{fig:CADPromptExample_1}
\end{figure*}

This subsection presents representative example from CADCodePrompt dataset to illustrate the natural language descriptions paired with their corresponding CAD code implementations (See Figure \ref{fig:CADPromptExample_1}). The example demonstrates how natural language of CAD design can be translated into executable Python code using CADQuery.

\begin{figure*}[t]
  \centering
  \includegraphics[height=6.5cm]{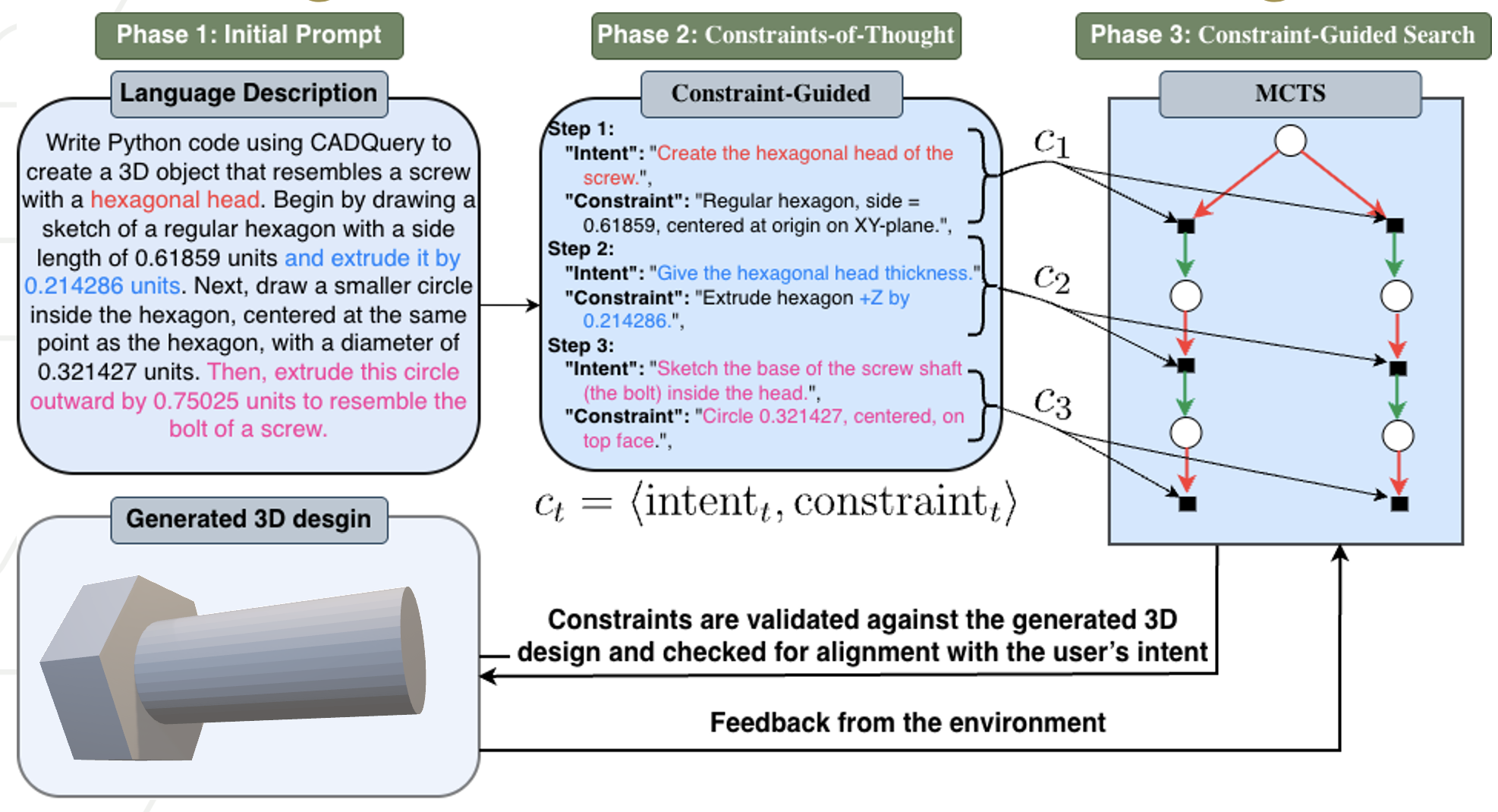}
  \caption{\textcolor{black}{Const-o-T empowers LLMs to (i) infer intent statements and (ii) extract executable constraints from high-level strategies, which restrict the admissible action set during search and act as search controllers rather than post-hoc output filters. These constraints guide a budgeted, constraint-guided variant of MCTS, steering exploration toward rule-compliant and high-quality actions while remaining robust to moderate constraint errors.}}
  \label{fig:schematicCAD}
\end{figure*}

\begin{figure*}[h]
\centering
\scriptsize
\begin{tcolorbox}[
    title=Constraints-of-Thought Prompt for Strategic Planning in the Risk Game,
    colback=gray!5,
    colframe=black,
    colbacktitle=gray!20,
    coltitle=black,
    sharp corners,
    width=.99\linewidth,
]

        You are a \textbf{strategic assistant for the Risk board game}. Your task is to \textbf{immediately allocate all troops} according to the Commander's intent, using step-by-step reasoning based on Constraints-of-Thought (Const-o-T) and Monte Carlo Tree Search (MCTS) decision-making. 

\vspace{1mm}
\textbf{Input Provided:}
\begin{itemize}
    \item  A natural language description of the Commander's intent for this turn's troop placement (e.g., "Fortify the Green continent and strengthen borders.").
    \item  The current map status (countries, ownership, and unoccupied territories).
\end{itemize}

\textbf{Your Output:}
\begin{itemize}
    \item Generate a sequential and complete plan for troop placement \textbf{for this turn only}.
\end{itemize}

For \textbf{each step}, provide:
\begin{itemize}
    \item A concise, natural-language \textbf{intent} (why the troop is being placed in that country)
    \item A \textbf{formal placement constraint} (e.g., ``Place 3 troops on Green\_C'')
    \item The exact country and number of troops as: \texttt{["Green\_C", 3]}
    \item The Placement must include a valid country and a numeric troop count; the troop count cannot be None, empty, or non-numeric. Only output a numeric value for the number of troops.

\end{itemize}

\textbf{Example Constraints:}
\begin{itemize}
    \item Place ‘n’ troops on Country ‘X’
    \item Attack Country ‘X’ from Country ‘Y’ with ‘n’ troops
    \item Move ‘n’ troops to Country ‘X’ from Country ‘Y’
    \item Add ‘n’ troops to reinforce Country 'X'
\end{itemize}

\vspace{2mm}
\textbf{Game Environment:}

Risk is a board game in which an army commander tries to take over the world by defeating all enemy troops and controlling all countries. Risk is a simplified version of real conflict, and has rules to reflect this:
\begin{itemize}
    \item Players control countries by having troops in them.
    \item The more countries and continents a player controls, the more resources they get.
    \item Players win countries from other players by battling with their troops.
    \item The more troops a player has when battling, the more likely they are to win.
    \item Players can only attack or be attacked by countries that are next to them.
    \item Some map connections are one-way only.
\end{itemize}

\vspace{1mm}
        Our modified RISK Map contains 5 continents - Red, Green, Purple, Yellow and Blue. Each continent is made up of countries. Red continent has 3 countries, Green has 5 countries, Purple has 5 countries, Yellow has 4 countries and Blue has 4 countries. Green\_A, Yellow\_B, Blue\_C, etc. are referred to as countries or territories Green, Yellow, Blue, Red, Purple are referred to as continents. Continents also have different connections between them through which the troops can move. These connections are one way i.e troops from the source country can only move to the destination country and not the other way round. The map has the following connections - Yellow\_D is connected to Green\_A, Greed\_D is connected to Red\_A, Red\_A is connected to Green\_D, Red\_B is connected to Purple\_E, Red\_C is connected to Yellow\_B, Red\_C is connected to Blue\_B, Blue\_A is connected to Yellow\_C, Yellow\_C is connected to Blue\_D, Blue\_C is connected to Purple\_A, Purple\_A is connected to Green\_E and Green\_E is connected to Purple\_A.

\vspace{1mm}
\textbf{INPUT}
\\
\textbf{Commander's Intent:} \verb|{Strategy_Description}|
\\
\textbf{The Map Status:} \verb|{mapStatus}|
\\
You may place troops only in countries that are unoccupied according to the current map status.

\vspace{2mm}
\textbf{OUTPUT FORMAT (JSON-like):}
\begin{verbatim}
Constraint-of-thoughts [
  {
    "step_id": 1,
    "intent": "Reinforce Green_C to protect the continent's border.",
    "constraint": "Place 5 troops on Green_C",
    "placement": ["Green_C", 5]
  }
]
\end{verbatim}

\vspace{2mm}
\textbf{INSTRUCTIONS}
\begin{itemize}
    \item Focus only on troop placement this turn. Do not suggest attacks, moves, or future planning.
    \item Use formal placement constraints for each action.
    \item Do not select the same country more than once.
    \item Be concise and ensure placements clearly support the Commander's current intent.
    \item Output the full CoT sequence in the specified format.
\end{itemize}

\vspace{2mm}
\textbf{Now, reason step-by-step and output your immediate troop placement sequence.}

\end{tcolorbox}
\caption{Prompt Template: Constraints-of-Thought for Risk Game Troop Placement}
\label{fig:Constraint}
\end{figure*}\begin{figure*}[h]
\centering
\scriptsize
\begin{tcolorbox}[
    title=Constraints-of-Thought Prompt for Strategic Planning in the Risk Game,
    colback=gray!5,
    colframe=black,
    colbacktitle=gray!20,
    coltitle=black,
    sharp corners,
    width=.99\linewidth,
]

        You are a \textbf{strategic assistant for the Risk board game}. Your task is to \textbf{immediately allocate all troops} according to the Commander's intent, using step-by-step reasoning based on Constraints-of-Thought (Const-o-T) and Monte Carlo Tree Search (MCTS) decision-making. 

\vspace{1mm}
\textbf{Input Provided:}
\begin{itemize}
    \item  A natural language description of the Commander's intent for this turn's troop placement (e.g., "Fortify the Green continent and strengthen borders.").
    \item  The current map status (countries, ownership, and unoccupied territories).
\end{itemize}

\textbf{Your Output:}
\begin{itemize}
    \item Generate a sequential and complete plan for troop placement \textbf{for this turn only}.
\end{itemize}

For \textbf{each step}, provide:
\begin{itemize}
    \item A concise, natural-language \textbf{intent} (why the troop is being placed in that country)
    \item A \textbf{formal placement constraint} (e.g., ``Place 3 troops on Green\_C'')
    \item The exact country and number of troops as: \texttt{["Green\_C", 3]}
    \item The Placement must include a valid country and a numeric troop count; the troop count cannot be None, empty, or non-numeric. Only output a numeric value for the number of troops.

\end{itemize}

\textbf{Example Constraints:}
\begin{itemize}
    \item Place ‘n’ troops on Country ‘X’
    \item Attack Country ‘X’ from Country ‘Y’ with ‘n’ troops
    \item Move ‘n’ troops to Country ‘X’ from Country ‘Y’
    \item Add ‘n’ troops to reinforce Country 'X'
\end{itemize}

\vspace{2mm}
\textbf{Game Environment:}

Risk is a board game in which an army commander tries to take over the world by defeating all enemy troops and controlling all countries. Risk is a simplified version of real conflict, and has rules to reflect this:
\begin{itemize}
    \item Players control countries by having troops in them.
    \item The more countries and continents a player controls, the more resources they get.
    \item Players win countries from other players by battling with their troops.
    \item The more troops a player has when battling, the more likely they are to win.
    \item Players can only attack or be attacked by countries that are next to them.
    \item Some map connections are one-way only.
\end{itemize}

\vspace{1mm}
        Our modified RISK Map contains 5 continents - Red, Green, Purple, Yellow and Blue. Each continent is made up of countries. Red continent has 3 countries, Green has 5 countries, Purple has 5 countries, Yellow has 4 countries and Blue has 4 countries. Green\_A, Yellow\_B, Blue\_C, etc. are referred to as countries or territories Green, Yellow, Blue, Red, Purple are referred to as continents. Continents also have different connections between them through which the troops can move. These connections are one way i.e troops from the source country can only move to the destination country and not the other way round. The map has the following connections - Yellow\_D is connected to Green\_A, Greed\_D is connected to Red\_A, Red\_A is connected to Green\_D, Red\_B is connected to Purple\_E, Red\_C is connected to Yellow\_B, Red\_C is connected to Blue\_B, Blue\_A is connected to Yellow\_C, Yellow\_C is connected to Blue\_D, Blue\_C is connected to Purple\_A, Purple\_A is connected to Green\_E and Green\_E is connected to Purple\_A.

\vspace{1mm}
\textbf{INPUT}
\\
\textbf{Commander's Intent:} \verb|{Strategy_Description}|
\\
\textbf{The Map Status:} \verb|{mapStatus}|
\\
You may place troops only in countries that are unoccupied according to the current map status.

\vspace{2mm}
\textbf{OUTPUT FORMAT (JSON-like):}
\begin{verbatim}
Constraint-of-thoughts [
  {
    "step_id": 1,
    "intent": "Reinforce Green_C to protect the continent's border.",
    "constraint": "Place 5 troops on Green_C",
    "placement": ["Green_C", 5]
  }
]
\end{verbatim}

\vspace{2mm}
\textbf{INSTRUCTIONS}
\begin{itemize}
    \item Focus only on troop placement this turn. Do not suggest attacks, moves, or future planning.
    \item Use formal placement constraints for each action.
    \item Do not select the same country more than once.
    \item Be concise and ensure placements clearly support the Commander's current intent.
    \item Output the full CoT sequence in the specified format.
\end{itemize}

\vspace{2mm}
\textbf{Now, reason step-by-step and output your immediate troop placement sequence.}

\end{tcolorbox}
\caption{Prompt Template: Constraints-of-Thought for Risk Game Troop Placement}
\label{fig:Constraint}
\end{figure*}

\begin{figure*}[h]
\centering
\scriptsize
\begin{tcolorbox}[
    title=Constraints-of-Thought Prompt for CAD code generation,
    colback=gray!5,
    colframe=black,
    colbacktitle=gray!20,
    coltitle=black,
    sharp corners,
    width=.99\linewidth,
]

\textbf{You are a symbolic CAD modeling assistant.} Your task is to generate a 3D CAD object by reasoning step-by-step using \textbf{Constraints-of-Thought (Const-o-T)} — a structured form of geometric planning based on user instructions.

\vspace{1mm}
You will be given:
\begin{itemize}
    \item A natural language description of a 3D object
\end{itemize}

\textbf{Your output must consist of a sequence of steps}, where each step includes:
\begin{enumerate}
    \item A short natural-language \textbf{intent} (what to do and why)
    \item A plain-English geometric \textbf{constraint} (e.g., ``The base should be a box with width 1.0, depth 0.75, and height 0.25'', or ``The hole should be centered on the base, with radius 0.1, and aligned along the Z-axis'')
\end{enumerate}

Each step must logically build upon previous ones.

\vspace{1mm}
\textbf{INPUT:}

\textbf{Natural Language Description:} \verb|{language_description}|

\vspace{1mm}
\textbf{Output format (JSON-like):}
\begin{verbatim}
{Constraints-of-Thought[
  {"Step_id": 1,
   "Intent": "Create the base plate.",
   "Constraint": "Make a rectangular box with width 1.0 units,
   depth 0.75 units, and height 0.25 units."},
  {"Step_id": 2,
   "Intent": "Add a centered hole.",
   "Constraint": "Drill a circular hole with radius 0.1 units at the center of the base,
   oriented along the Z-axis."}
]}
\end{verbatim}

\vspace{1mm}
\textbf{INSTRUCTIONS:}
\begin{itemize}
    \item Think like a constraint solver: extract \textbf{clear symbolic relationships} but describe them in English.
    \item Use \textbf{plain sentences} that can be easily mapped to CAD operations.
    \item Be concise. Avoid unnecessary primitives.
    \item The final object must follow all described constraints.
\end{itemize}

\vspace{1mm}
\textbf{Now reason step-by-step and output the full Constraints-of-Thought (Const-o-T) sequence.}

\end{tcolorbox}
\caption{Prompt for symbolic 3D CAD modeling using Constraints-of-Thought (Const-o-T).}
\label{fig:risk_prompt}
\end{figure*}

\begin{figure*}[h]
\centering
\scriptsize
\begin{tcolorbox}[
    title=Evaluation Prompt for Math Arithmetic Step,
    colback=gray!5,
    colframe=black,
    colbacktitle=gray!20,
    coltitle=black,
    sharp corners,
    width=.99\linewidth,
]

You are an evaluation agent tasked with assessing whether a reasoning step effectively contributes to solving a math problem.

\vspace{1mm}
Your job is to evaluate a single step in the problem-solving process and return a score between 0 and 1, where:
\begin{itemize}
    \item - 1 means the step is highly useful and directly helps solve the problem,
    \item - 0 means the step is irrelevant, misleading, or incorrect,
    \item - Intermediate values (e.g., 0.5) indicate partial usefulness or vague contribution.
\end{itemize}

\vspace{1mm}
\textbf{INPUT:}
\begin{itemize}
    \item \textbf{Problem:} \verb|{question}|
    \item \textbf{Step to evaluate:} \verb|{Step}|
\end{itemize}

\vspace{1mm}
\textbf{Output format (JSON):}
\begin{verbatim}
{"score": float between 0 and 1}
\end{verbatim}

\vspace{1mm}
\textbf{Now assess the step and output the score.}

\end{tcolorbox}
\caption{Prompt for evaluating math arithmetic step in MCTS.}
\label{fig:math_eval_prompt}
\end{figure*}

\begin{figure*}[h]
\centering
\scriptsize
\begin{tcolorbox}[
    title=Evaluation Prompt for CAD code Step,
    colback=gray!5,
    colframe=black,
    colbacktitle=gray!20,
    coltitle=black,
    sharp corners,
    width=.99\linewidth,
]

You are an evaluation agent tasked with assessing whether a **generated CAD modeling code snippet** effectively contributes to building the intended 3D model.

\vspace{1mm}
Your job is to evaluate a single step of **CAD code** within the Constraints-of-Thought pipeline and return a score between 0 and 1, where:
    \begin{itemize}
    \item - 1 means the code is highly useful, precise, and directly implements the intended modeling operation,
    \item - 0 means the code is irrelevant, incorrect, or harmful to the design,
    \item - Intermediate values (e.g., 0.5) indicate partial correctness, vague implementation, or incomplete alignment with the intended step.
\end{itemize}
    
\vspace{1mm}
\textbf{INPUT:}
\begin{itemize}
    \item \textbf{CAD Object Description:} \verb|{question}|
    \item \textbf{CAD Code to evaluate:} \verb|{Step}|
\end{itemize}

\vspace{1mm}
\textbf{Output format (JSON):}
\begin{verbatim}
{"score": float between 0 and 1}
\end{verbatim}

\vspace{1mm}
\textbf{Now assess the step and output the score.}

\end{tcolorbox}
\caption{Prompt for evaluating CAD code step in MCTS.}
\label{fig:CAD_eval_prompt}
\end{figure*}

\begin{figure*}[h]
\centering
\scriptsize
\begin{tcolorbox}[
    title=Constraints-of-Thought Prompt for math arithmetic,
    colback=gray!5,
    colframe=black,
    colbacktitle=gray!20,
    coltitle=black,
    sharp corners,
    width=.99\linewidth,
]

\textbf{You are a symbolic math solver.} Your task is to solve a math word problem by reasoning step-by-step using \textbf{Constraints-of-Thought (Const-o-T)} — a structured form of logical planning.

\vspace{1mm}
You will be given:
\begin{itemize}
    \item A natural language word problem involving arithmetic reasoning.
\end{itemize}

\textbf{Your output must consist of a sequence of steps}, where each step includes:
\begin{enumerate}
    \item A short natural-language \textbf{Intent} (what is being computed and why).
    \item A precise \textbf{Constraint} using symbolic math expressions (e.g., \texttt{x = 2 * y}, \texttt{total = cost\_per\_egg * eggs\_sold}).
\end{enumerate}

\vspace{1mm}
\textbf{INPUT:}

\textbf{Word Problem:} \verb|{question}|

\vspace{1mm}
\textbf{OUTPUT FORMAT (JSON-like):}
\begin{verbatim}
json{Constraint-of-thoughts [
  {
    "Step_id": 1,
    "Intent": "Determine the number of bolts of blue fiber needed for one robe.",
    "Constraint": "blue_bolts = 2"
  },
  {
    "Step_id": 2,
    "Intent": "Calculate the number of bolts of white fiber needed, 
    which is half as much as blue fiber.",
    "Constraint": "white_bolts = blue_bolts / 2"
  },
  {
    "Step_id": 3,
    "Intent": "Add the number of blue and white bolts to find
    the total number of bolts needed.",
    "Constraint": "total_bolts = blue_bolts + white_bolts"
  }
]}
\end{verbatim}

\vspace{1mm}
\textbf{INSTRUCTIONS:}
\begin{itemize}
    \item Use precise arithmetic operations for each step (addition, subtraction, multiplication, division).
    \item Define and name variables explicitly to reflect the quantities being calculated (e.g., \texttt{total\_bolts}, \texttt{num\_apples}, \texttt{cost\_per\_item}).
    \item Each step should logically follow from the previous one, building upon the calculations.
    \item Only use the information given in the problem — no assumptions or outside knowledge.
    \item Your solution must contain \textbf{no more than 7 steps}. Merge or skip trivial operations when appropriate.
\end{itemize}

\end{tcolorbox}
\caption{Prompt for math arithmetic using Constraints-of-Thought (Const-o-T).}
\label{fig:math_prompt}
\end{figure*}



\end{document}